\documentclass[lettersize,journal]{IEEEtran}
\usepackage{amsmath,amsfonts}
\usepackage{array}
\usepackage{textcomp}
\usepackage{graphicx}
\usepackage{cite}
\usepackage{booktabs}
\usepackage{multirow}
\usepackage{xcolor}
\usepackage{url}
\usepackage[hidelinks]{hyperref}
\usepackage{algorithm}
\newcommand{\fps}{\textsc{fps}}
\newcommand{\dreamlite}{DreamLite-mobile}
\newcommand{\qwen}{Qwen3-VL}

\begin{document}

\title{Inverting the Streaming-Diffusion Bottleneck:\\
Video-Rate MLLM-Conditioned Edit Diffusion\\
on a Consumer GPU}

\author{Yoshiyuki Ootani,~\IEEEmembership{Independent Researcher}
\thanks{Y.~Ootani is an independent researcher. Email: info@ootanl.com.
ORCID: 0009-0004-4432-0660.}%
\thanks{Source code, evaluation harness, and the released v3 LLLite
adapter are at \protect\url{https://github.com/otanl/dreamlite-stream};
the release accompanying this preprint is archived at
\protect\url{https://doi.org/10.5281/zenodo.20389428}.}%
\thanks{Preprint, May 2026. Under review.}}

\maketitle

\begin{abstract}
Aggressive distillation of the diffusion U-Net inverts the per-frame
bottleneck of real-time text-to-image pipelines: once the denoiser
is a 4-step or 1-step distilled student, the text encoder becomes
the critical path. This inversion is most acute in vision-aware
edit diffusion, where the encoder is a multimodal large language
model (MLLM). We study the case of a $0.39$\,B distilled edit
U-Net paired with a $2.13$\,B MLLM text encoder (Qwen3-VL) and
present a streaming pipeline targeted at this regime, addressing the
inversion with three mechanisms that keep the encoder off the
denoiser's critical path rather than trying to shrink the encoder
itself: asymmetric side-stream / main-stream CUDA pipelining with
batched text-encoder amortisation (and optional static-prompt
caching), a compile-friendly ControlNet-LLLite reformulation that
folds the entire U-Net + adapter stack into a single fused graph, and
a periodic conditioning-refresh schedule with a hook subset that
amortises the per-frame conditioning cost. On a single consumer
RTX 3090 Ti at $512\!\times\!512$ this sustains $27$--$30$\,\fps{}
over a 480-frame run; at the same operating point steady-state
throughput scales to $55$\,\fps{} on RTX 4090 and $74$\,\fps{} on
RTX 5090 (full operating points in Table~\ref{tab:main}). This
demonstrates
that once denoiser distillation is aggressive enough, further
per-frame gains come from encoder-side systems work rather than from
compressing the denoiser further -- the opposite lever from the one
the streaming-diffusion literature has optimised to date. We report
video-rate streaming \emph{throughput} rather than interactive low
latency, and locate our numbers against same-stack StreamDiffusion
re-runs as systems context, not as a benchmark superiority claim.
For the trained \emph{oil-painting} style, the released temporal
adapter generalises within in-clip noise to 19 unused
DAVIS-2017 sequences and 15 non-DAVIS clips from seven sources;
prompt-level generalisation to unseen style families is bounded
and reported separately.
\end{abstract}


\section{Introduction}\label{sec:intro}

\IEEEPARstart{R}{eal-time} streaming video translation with diffusion
models has been driven by text-conditioned image diffusion stacks in
which the U-Net denoiser dominates the per-frame critical path and the
text encoder is a CLIP-class network of $\sim$125\,M parameters. Once
the denoiser is aggressively distilled (a 4-step or 1-step student
running under \texttt{torch.compile} or TensorRT), this familiar
bottleneck inverts: the text encoder becomes the per-frame critical
path, and the systems techniques developed for the denoiser-dominated
regime stop being the right design centre.

\label{sec:intro:generality}
We name this operating point the
\emph{small-distilled-denoiser plus heavy-MLLM-encoder} (SDD-MTE)
configuration: a $\leq$$1$\,B distilled diffusion denoiser running
under \texttt{torch.compile} or TensorRT, paired with a
$\geq$$2$\,B vision-aware multimodal language model that jointly
attends to a source image and an instruction string. We are careful
not to overclaim here: our evidence is a single instantiation
(\dreamlite{}+\qwen{}), so we present SDD-MTE as a characterised
operating point and a \emph{hypothesis} about a class, not a
demonstrated regime. Whether the three mechanisms transfer to other
distilled-denoiser/MLLM-encoder pairings is an open question we
cannot settle with $n{=}1$. What \emph{does} carry beyond this
configuration is a methodological caution we quantify in this regime:
warping error~\cite{lai2018temporal} is minimised by high-frequency
loss, so a smoother (blurrier) output scores \emph{better} on it. The
failure mode is known in principle; our contribution is to measure how
large it is for a distilled streaming styliser and to show, against an
independent baseline (\S\ref{sec:exp:throughput}), that warping error
must be read together with a spectral/edge probe rather than alone.

The most acute case of this inversion arises in vision-aware edit-mode
diffusion: models that consume both a source image and a textual
instruction in order to produce an instruction-conditioned re-render.
This class has converged on multimodal large language model (MLLM)
text encoders that jointly process image and text tokens.
\dreamlite{}~\cite{feng2026dreamlite} is representative — a $0.39$\,B
parameter U-Net, distilled to 4 steps, conditioned on \qwen{}-2B
\cite{qwen3vl2025} (2.13\,B parameters) which jointly attends to a
$256\!\times\!256$ source patch and an instruction string. The text
encoder is roughly $5\!\times$ larger than the denoiser by parameter
count, and it carries an intrinsic CPU--GPU synchronisation point
inside its rotary position-embedding code that prevents simple
single-stream dispatch from overlapping it with U-Net compute.

The temporal-consistency adapters that emerged for SD-class
denoisers, in particular ControlNet-LLLite~\cite{kohya2023lllite},
make this regime harder rather than easier. Their forward path uses
Python branches and tensor-identity tests that break
\texttt{torch.compile}'s graph capture, so a naive LLLite-augmented
U-Net runs $\sim$3.5$\times$ slower than the base U-Net even at the
same arithmetic intensity. Wide-range MLLM TE outputs (std$=37$,
max$=1160$ in our profiling) also overflow FP16 attention, so the
TensorRT path that StreamDiffusion-class systems
\cite{kodaira2023streamdiffusion} rely on for headline throughput is
unavailable here.

The root cause of the inversion is architectural, not incidental: an
MLLM text encoder carries an intrinsic CPU--GPU synchronisation point
and Python-level control flow that a CLIP-class encoder does not, so
it cannot simply be dispatched on the same stream as a compiled U-Net
without stalling it. Fixing this requires treating the encoder as a
first-class latency-budget item -- something to pipeline, amortise,
and compile around -- rather than an afterthought behind a fast
denoiser. We address this cause with three mechanisms that generalise
beyond the specific backbone pair we instantiate them on
(\S\ref{sec:intro:generality}), each with a directly measured
per-frame impact:

\begin{enumerate}
\item \emph{Asymmetric side-stream / main-stream pipelining}
  (\S\ref{sec:method:pipeline}) places the MLLM TE on a dedicated
  CUDA stream and runs it concurrently with the compiled U-Net on the
  main stream, hiding the TE's intrinsic \texttt{.item()} sync inside
  the U-Net step. Two opt-in static-prompt approximations are
  exposed for streams where the user instruction is fixed and the
  source visual content changes slowly --- a single-frame TE
  prompt (\texttt{te\_batch\_one}) and an $N$-batch TE refresh
  schedule (\texttt{te\_refresh\_every}) --- which amortise the
  TE cost further but are not used for any reported number unless
  explicitly stated.
\item A \emph{compile-friendly LLLite reformulation}
  (\S\ref{sec:method:lllite}) removes Python branches and
  tensor-identity changes from the 108-hook adapter forward, allowing
  \texttt{torch.compile} to fold the entire U-Net + adapter stack
  into a single fused graph.
\item A \emph{periodic conditioning-refresh and hook-subset schedule}
  (\S\ref{sec:method:cond}) amortises the LLLite cond-image encoder
  cost across $N$ batches, exposes a smoothing artifact in which
  warping-error is mechanically lowered by output blur, and threads
  the per-frame Farneback flow asynchronously during prefetch.
\end{enumerate}

On a single consumer RTX~3090~Ti at $512\!\times\!512$ the resulting
pipeline sustains $27.4$\,\fps{} over a $480$-frame run at batch
$B{=}8$ with end-to-end p50 latency $\approx$$0.5$\,s, and
$29.6$\,\fps{} at $B{=}16$. The same operating point measures
$54.9$\,\fps{} on RTX~4090 and $74.1$\,\fps{} on RTX~5090. We do not
claim superiority over StreamDiffusion-class systems and instead
locate our numbers against a same-stack rerun of their reference
configuration as systems context (\S\ref{sec:exp:throughput}). The
released temporal adapter is evaluated in-domain on DAVIS-2017 and
out-of-domain on 19 unused DAVIS sequences plus a 15-clip
cross-dataset sanity check spanning seven non-DAVIS sources
(\S\ref{sec:exp:transfer}).

\emph{Paper organisation.} Section~\ref{sec:related} surveys
streaming-V2V, MLLM-conditioned generation, and temporal-consistency
work. Section~\ref{sec:method} details the three engineering
mechanisms. Section~\ref{sec:exp} reports sustained throughput,
temporal quality, transfer evaluation, and hardware scaling.
Section~\ref{sec:neg} documents five negative results and a
distillation case study. Section~\ref{sec:conclusion} closes with
deployment implications.

\section{Related Work}\label{sec:related}

\subsection{Streaming and Real-Time Diffusion Video Pipelines}
StreamDiffusion~\cite{kodaira2023streamdiffusion} establishes the
modern reference point for real-time text-conditioned image
diffusion: batched-timestep dispatch, residual classifier-free
guidance, stochastic similarity filtering, and TensorRT compilation
together reach 91\,\fps{} on SD-Turbo and 38\,\fps{} on Kohaku v2.1
$+$ LCM-LoRA at $512\!\times\!512$ on an RTX~4090. StreamV2V
\cite{liang2025streamv2v} extends the framework to video-to-video by
maintaining a cross-frame feature bank; StreamDiffusionV2
\cite{feng2025streamdiffusionv2} adds SLO-aware batching and a
sink-token--guided rolling KV cache for multi-GPU video-diffusion
backbones; StreamDiT~\cite{kodaira2025streamdit} and MotionStream
\cite{shin2026motionstream} target streaming text-to-video
\emph{generation} on dedicated DiT backbones; RTR-DiT
\cite{lyu2026rtrdit} and Live2Diff~\cite{xing2024live2diff} pursue
causal real-time video translation with autoregressive DiT or
uni-directional 3D-attention bases respectively. A concurrent line
targets streaming \emph{edit} settings on DiT substrates:
SANA-Streaming~\cite{zhao2026sanastreaming} pairs a hybrid
full-/linear-attention DiT with a \qwen{}-class MLLM text encoder
and a 4-step distillation, demonstrating real-time streaming video
editing on a higher-end GPU at $1280\!\times\!704$;
Stream-R1~\cite{wu2026streamr1} introduces reward-distillation for
autoregressive streaming video generation; and the \texttt{scope}
system~\cite{fosdick2026vacestreaming} adapts VACE control to a
real-time autoregressive video-diffusion path without retraining.
Streaming Video
Diffusion~\cite{chen2024streamingvdiff} formalises the
online-editing setting with a video-token cache; Denoising Reuse
\cite{wang2025denoisingreuse} amortises noise-prediction cost across
frames by reusing inter-frame motion structure. ParaDiGMS
\cite{shih2023parallel}, DistriFusion~\cite{li2024distrifusion}, and
PipeFusion~\cite{fang2025pipefusion} parallelise the denoising
trajectory across timesteps or GPUs for offline throughput. A
recent line of caching / reuse / pruning work on the encoder side is
directly relevant: VLCache~\cite{qin2025vlcache} reuses VLM encoder
state across repeated multimodal inputs,
Skip-Vision~\cite{zeng2025skipvision} accelerates VLM inference via
adaptive visual-token skipping, InfiniPot-V~\cite{kim2025infinipotv}
compresses KV cache for streaming video understanding under memory
bounds, HAWK~\cite{zhu2026hawk} prunes visual tokens by head-importance
on \qwen{}-class encoders with reported large visual-token-budget
reductions at near-parity downstream accuracy, and
IDPruner~\cite{tan2026idpruner} jointly preserves importance and
diversity across visual tokens for multi-MLLM-family pruning. Adaptive
caching has also emerged on the denoiser side: AdaCache
\cite{kahatapitiya2025adacache} adapts cache cadence to motion in
diffusion transformers, and Elastic-Cache
\cite{nguyentri2025elasticcache} formalises when/where to refresh
in diffusion LLMs. We adopt a deliberately simple periodic-refresh
policy and discuss adaptive alternatives in
\S\ref{sec:exp:quality} and \S\ref{sec:conclusion}. Our setting
differs from all of the above in two ways: a vision-aware MLLM
text encoder makes the per-frame critical path the encoder rather
than the denoiser, and a single consumer GPU forbids the multi-GPU
patch parallelism that the offline systems assume.
Table~\ref{tab:streamv2v-landscape} positions our work along base
model, conditioning modality, and per-frame critical path.

\begin{table*}[t]
\caption{Streaming video-to-video systems by base model and
conditioning modality. Ours occupies the distilled-edit-UNet +
MLLM-TE cell with the per-frame critical path on the encoder rather
than the denoiser.}
\label{tab:streamv2v-landscape}
\centering
\footnotesize
\begin{tabular}{@{}lllll@{}}
\toprule
\textbf{Work} & \textbf{Base model} & \textbf{Conditioning} & \textbf{Temporal mechanism} & \textbf{Per-frame critical path} \\
\midrule
StreamDiffusion~\cite{kodaira2023streamdiffusion} & SD / SD-Turbo image diff. & text & per-frame i2i (none) & U-Net \\
StreamV2V~\cite{liang2025streamv2v}               & SD image diff. & text & cross-frame feature bank & U-Net \\
Live2Diff~\cite{xing2024live2diff}                & video-diffusion (3D) & text & uni-directional temporal attn. & video-diff.\ denoiser \\
StreamDiffusionV2~\cite{feng2025streamdiffusionv2}& video-diff.\ 1.3--14B & text & rolling KV cache + SLO sched. & video-diff.\ denoiser \\
StreamDiT~\cite{kodaira2025streamdit}             & streaming DiT 4B (gen.) & text & moving frame buffer + window attn. & DiT denoiser \\
MotionStream~\cite{shin2026motionstream}          & causal video DiT (gen.) & text + motion traj. & sliding window + attn.\ sinks & video DiT denoiser \\
RTR-DiT~\cite{lyu2026rtrdit}                      & autoregressive DiT (video) & text / reference & AR KV-cache rerender & video DiT denoiser \\
SANA-Streaming~\cite{zhao2026sanastreaming}       & hybrid DiT (edit) + MLLM TE & image + instruction & per-frame edit + DiT cache & DiT denoiser + MLLM TE \\
Stream-R1~\cite{wu2026streamr1}                   & AR streaming video DiT & text & reward-distilled AR rollout & video DiT denoiser \\
\texttt{scope}~\cite{fosdick2026vacestreaming}    & VACE (AR video-diff.) & VACE control signals & real-time control adaptation & VACE control $+$ denoiser \\
\midrule
\textbf{Ours} & \textbf{distilled edit U-Net + MLLM TE} & \textbf{image + instruction} & \textbf{LLLite temporal adapter} & \textbf{MLLM TE} \\
\bottomrule
\end{tabular}
\end{table*}

\paragraph{Regime contrast vs.\ concurrent DiT streaming systems.}
The closest concurrent system in spirit,
SANA-Streaming~\cite{zhao2026sanastreaming}, validates the
MLLM-text-encoder direction on a hybrid full/linear-attention DiT
trained end-to-end with a \qwen{}-class encoder. Our regime is
intentionally complementary along three axes: (i) the substrate is a
publicly-distilled edit U-Net rather than an end-to-end-trained DiT;
(ii) the operating budget is a single consumer GPU rather than a
higher-end Blackwell-class card; and (iii) we do not attempt
cross-base text-encoder swap on existing distilled DiT bases, because
in preliminary work the text-encoder representation in those bases is
tightly entangled with the rest of the network and a post-hoc swap
does not converge in our setting. SANA-Streaming's trained-together
strategy is, to our knowledge, the only existing demonstration of
``distilled DiT $+$ MLLM TE'' at video rate.

\subsection{Vision-Aware and MLLM-Conditioned Generation}
A separate strand pairs diffusion bases with multimodal language
models for instruction-conditioned editing or generation. GPT4Video
\cite{wang2024gpt4video} couples a unified MLLM with diffusion
decoders for instruction-followed multimodal generation; MoTrans
\cite{li2024motrans} customises motion via text-driven video
diffusion guidance; unified MLLM-edit systems such as BAGEL
\cite{deng2025bagel} (7\,B active) and OmniGen2~\cite{wu2025omnigen2}
(4\,B diff.\ + 3\,B VLM) place a full MLLM in the conditioning path.
\dreamlite{}~\cite{feng2026dreamlite} compresses this class with a
$0.39$\,B distilled U-Net and a $2.13$\,B \qwen{}
\cite{qwen3vl2025} encoder. Among documented vision-aware distilled
edit models, this pairing is the smallest combination that retains
instruction-following at video-stylization quality; the asymmetry it
creates between a fast denoiser and a heavy encoder is exactly the
regime this paper targets.

\subsection{Temporal Consistency and Adapter-Based Conditioning}
Cross-frame attention and 3D convolution define the heavyweight
route to temporal consistency (AnimateDiff~\cite{guo2023animatediff},
Stable Video Diffusion~\cite{blattmann2023stable}). Zero-shot routes
rely on optical flow or keyframe re-rendering (Rerender-A-Video
\cite{yang2023rerender}, FlowVid~\cite{liang2024flowvid}); both are
offline. Among recent temporal-consistent editing systems, TCVE~\cite{wang2024tcve}
imposes temporal-consistent video editing via image-diffusion bases
with a dedicated motion module; spatio-temporal energy-guided
diffusion~\cite{yang2025stegd} unifies zero-shot synthesis and
editing with an explicit energy prior; UniVST~\cite{song2024univst}
targets training-free localised video style transfer; TVG
\cite{zhang2025tvg} formalises a training-free transition-video
generator; FluencyVE~\cite{cai2026fluencyve} replaces full
temporal attention with a Mamba state-space mixer plus bypass
attention for editing efficiency. ControlNet~\cite{zhang2023controlnet} and ControlNet-LLLite
\cite{kohya2023lllite} provide the adapter mechanism we build on:
LLLite uses per-attention zero-initialised residuals
($\sim$13\,M parameters at the default rank) gated by a per-block
CNN encoder of a conditioning image. We re-purpose this adapter from
\emph{spatial} conditioning (canny, depth, pose) to a \emph{temporal}
signal, the warped previous decoded frame, with the teacher being
the post-hoc-blended base output at $\alpha{=}0.85$.

\subsection{Distillation and Step Reduction}
Latent Consistency Models~\cite{luo2023lcm}, adversarial
distillation~\cite{sauer2024adversarial}, and progressive
adversarial distillation~\cite{lin2024sdxllightning} reduce inference
step count by retraining. We \emph{do not} retrain in this paper.
Instead we show empirically (\S\ref{sec:exp:throughput}) that the
4-step distilled \dreamlite{} base degrades gracefully to $K{=}1$
inference on the trained style prompts at a measurable
teacher-fidelity cost, providing a useful operating point for
deployment without an additional distillation pass.

\section{Method}\label{sec:method}

\subsection{Notation and Setup}
\label{sec:method:setup}
\dreamlite{}~\cite{feng2026dreamlite} pairs a $0.39$\,B U-Net (TAESDXL
VAE; FlowMatchEuler scheduler~\cite{esser2024scaling}) with \qwen{}
\cite{qwen3vl2025} as the text encoder; text-to-image
(\textsc{generate}) and image-edit (\textsc{edit}) modes share
weights. \textsc{edit} concatenates the noisy latent with an encoded
reference-image latent along the spatial-width axis, and the
TE prompt template injects a $\langle\texttt{image\_pad}\rangle$
visual token processed by \qwen{}'s vision tower. The released
checkpoint is 4-step distilled.

\paragraph{Denoising update} \dreamlite{} uses a FlowMatchEuler
schedule~\cite{esser2024scaling} over $K$ steps at noise levels
$\sigma_0{=}1 > \sigma_1 > \dots > \sigma_K{=}0$. Given latent $z$,
TE hidden state $h$, and LLLite conditioning $c$, one step is
\begin{equation}
z^{(k+1)} = z^{(k)} + (\sigma_{k+1}-\sigma_k)\,
  v_\theta\!\big(z^{(k)}, \sigma_k;\, h, c\big),
\label{eq:flowstep}
\end{equation}
where $v_\theta$ is the U-Net's predicted velocity. $K{=}4$ is the
released (teacher) step count; $K{=}1$ collapses
Eq.~\eqref{eq:flowstep} to a single $\sigma{:}1\!\to\!0$ update
(\S\ref{sec:method:pipeline}).

\paragraph{LLLite adapter} ControlNet-LLLite~\cite{kohya2023lllite}
attaches $108$ adapter modules to the \textsc{LLLite}-eligible Linear
layers in U-Net attention blocks ($\textsc{attn1}.\{q,k,v\}$ and
$\textsc{attn2}.q$). For host Linear layer $i$ with input $x_i$, each
module computes a per-block CNN encoding $g_i$ of the conditioning
image $c$ and adds a zero-initialised residual to the layer output:
\begin{equation}
\mathrm{Linear}_i^{\text{LLLite}}(x_i, c) =
  \mathrm{Linear}_i(x_i) + \delta_i, \quad
  \delta_i = g_i(c)\cdot x_i,
\label{eq:lllite}
\end{equation}
with $g_i$ zero-initialised at the start of training so
$\delta_i{=}0$ initially. We train this adapter with the
Farneb\"ack-warped~\cite{farneback2003twoframe} previous decoded
frame as $c$ and the post-hoc-blended ($\alpha{=}0.85$) base output
as the teacher target.

\paragraph{Warping error} Temporal consistency is measured with the
warping error $\varepsilon_w$~\cite{lai2018temporal}: given
consecutive output frames $y_{n-1}, y_n$ and the Farneb\"ack optical
flow $\mathrm{flow}_{n-1\to n}$ between the corresponding
\emph{source} frames,
\begin{equation}
\varepsilon_w(n) = \big\| y_n - \mathrm{warp}(y_{n-1},
  \mathrm{flow}_{n-1\to n}) \big\|_2^2 ,
\label{eq:ew}
\end{equation}
averaged over a clip; lower is smoother. Because
Eq.~\eqref{eq:ew} is minimised by any transformation that flattens
high-frequency detail, we report it alongside a smoothing-collapse
check (\S\ref{sec:exp:transfer}) rather than in isolation.

All experiments use Python~3.10.0,
PyTorch~2.6.0$+$cu124, Triton~3.2 (Windows port), on a single RTX
3090~Ti unless noted otherwise.

\subsection{Asymmetric Side-Stream / Main-Stream Pipelining}
\label{sec:method:pipeline}

The reference implementation runs a \textsc{generate}-mode encode
followed by a 4-step denoise on a single Python thread. We
restructure it into four cumulative steps.

\paragraph{\texttt{torch.compile} of the U-Net} We compile only the
U-Net (\texttt{reduce-overhead}, \texttt{fullgraph=False},
\texttt{dynamic=False}). At fixed $512\!\times\!512$ shapes the
compiled U-Net is $3.1\times$ faster than eager.

\paragraph{CUDA-stream pipelining of TE and VAE-encode} The
\textsc{edit}-mode TE output depends on both the source image and
the prompt and in the base operating mode must be recomputed every
batch. Its $\sim$200\,ms latency at $B{=}1$ can however be hidden
behind the next frame's denoise. We launch TE and VAE-encode for
batch $n{+}1$ on a side CUDA stream while the default stream runs
batch $n$'s denoise and VAE-decode; synchronisation uses a single
CUDA event, so the two streams contend only at memory bandwidth.
The TE additionally carries an intrinsic \texttt{.item()}
synchronisation point inside its rotary position-embedding code;
placing the encoder on its own stream lets this stall overlap
U-Net compute on the main stream rather than serialising into the
per-frame critical path.

\paragraph{Zero-retraining $K{=}1$ inference} Running the released
4-step distilled weights at $K{=}1$ (a single $\sigma{=}1\!\to\!0$
flow-matching update) preserves recognisable stylisation at the same
prompt with $\varepsilon_w$ within 5\% of the $K{=}2$ trajectory's;
LPIPS to the $K{=}4$ teacher rises to $\approx$$0.40$
(Table~\ref{tab:main}). We trade this controlled fidelity loss for a
$2\times$ throughput gain; this is an inference-time operating point,
not a new distillation.

\paragraph{Asymmetric batched dispatch with TE caching} At fixed
$K{=}1$, Table~\ref{tab:batch} (\S\ref{sec:exp:throughput}) shows
that the \qwen{} TE amortises substantially better with batch than
the compiled U-Net: TE wall grows from $125$\,ms at $B{=}1$ to
$153$\,ms at $B{=}4$ and $233$\,ms at $B{=}8$ (sub-linearity
$s{=}0.306$ and $0.233$), while the compiled U-Net grows from
$23$\,ms to $74$\,ms and $142$\,ms ($s{=}0.804$ and $0.772$).
We exploit this asymmetry by buffering
$B{=}8$ frames before each pipeline call; per-frame TE drops from
$125$ to $\sim$$30$\,ms while per-frame U-Net cost rises only
marginally. Latency cost is $B{-}1{=}7$ frames of buffering
($\sim$$230$\,ms at $30$\,\fps{} source).

\begin{figure*}[t]
\centering
\includegraphics[width=\textwidth]{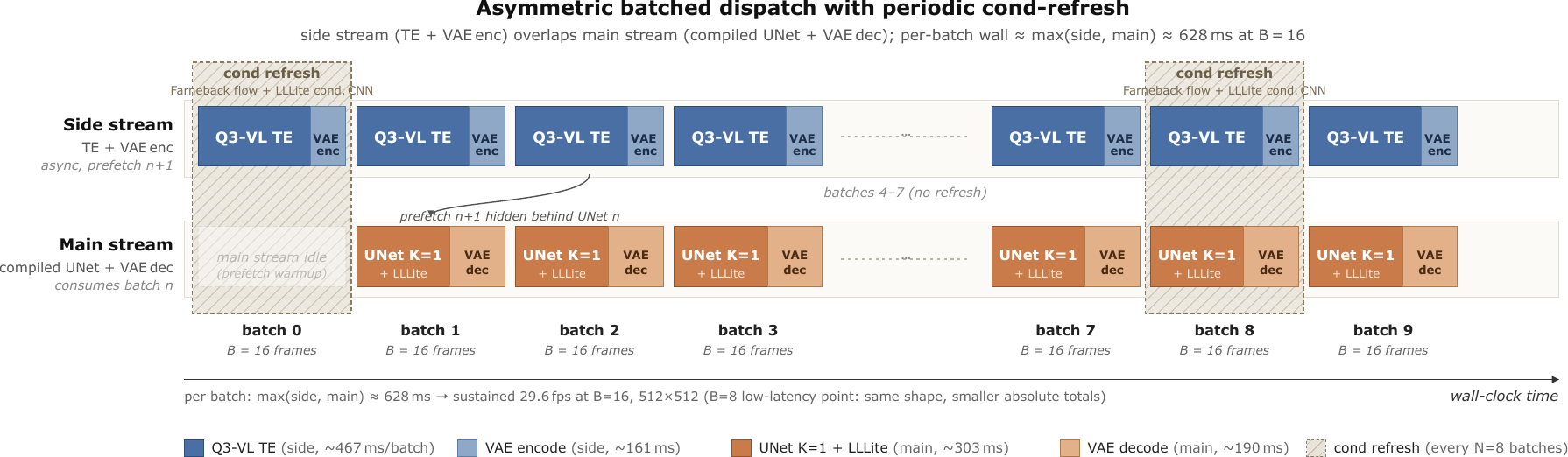}
\caption{Asymmetric batched dispatch on the \dreamlite{}+\qwen{}
stack (\textbf{unoptimised cond-refresh path}, matching
Table~\ref{tab:profile}). The \qwen{} TE and VAE-encode for batch
$n{+}1$ run on a CUDA \emph{side stream} while the compiled U-Net
($K{=}1$, \texttt{down\_blocks} LLLite hook subset) and VAE-decode
for batch $n$ run on the \emph{main stream}. Because the TE
dominates the side total ($\approx$$467$\,ms vs.\ $\approx$$303$\,ms
U-Net at $B{=}16$, Table~\ref{tab:profile}), the per-batch wall is
bounded by the side stream and the main stream is fully hidden
behind the TE except for the first batch's prefetch warm-up.
Hatched columns mark the periodic LLLite cond-refresh
(\S\ref{sec:method:cond}), fired every $N{=}8$ batches. The
$628$\,ms per-batch wall annotated here corresponds to the
$26.4$\,fps baseline row of Table~\ref{tab:sustained}
($16/0.628\!\approx\!25.5$\,fps in isolation; the additional gap
to $26.4$\,fps reflects the seven non-refresh batches in each
window of eight); threading the per-frame Farneb\"ack flow and
async-kicking it inside \texttt{prefetch\_batch} so it overlaps
the next side-stream slot is what reaches the $29.6$\,fps
sustained champion throughput reported in Table~\ref{tab:sustained}
and the body text.}
\label{fig:pipeline}
\end{figure*}

\paragraph{Optional TE caching modes for static-prompt streams}
When the user instruction is fixed across the stream and the
source visual content changes slowly, we additionally expose two
opt-in switches that further amortise the TE: (i)
\texttt{te\_batch\_one} encodes only one representative
$(\text{image}, \text{prompt})$ pair in the batch and broadcasts
its key/value tensors to the remaining $B{-}1$ frames, shrinking
the multimodal sequence by $B\times$; (ii)
\texttt{te\_refresh\_every}~$=N_{\text{TE}}$ caches the broadcast
TE output across $N_{\text{TE}}$ consecutive batches so it is
recomputed once per $B \cdot N_{\text{TE}}$ source frames. Both
modes are explicit approximations: $(i)$ accepts that the visual
content represented in the TE drifts from the actual per-frame
input across the batch, $(ii)$ extends that drift across the
cache window. They are independent of the per-batch LLLite
\emph{cond}-refresh schedule (\S\ref{sec:method:cond}), which
amortises the LLLite conditioning encoder rather than the
\qwen{} TE. Algorithm~\ref{alg:loop} writes out the resulting
per-batch loop, and Fig.~\ref{fig:pipeline} sketches the
two-stream schedule.

\begin{algorithm}[t]
\caption{Per-batch streaming inference loop ($K{=}1$ instance of
Eq.~\eqref{eq:flowstep} with LLLite conditioning from
Eq.~\eqref{eq:lllite}). Side stream and
main stream are dispatched concurrently and synchronised via a
single CUDA event. $\mathrm{TE}$ is the \qwen{} text encoder;
$\mathrm{LLLite.cond}(c) = (g_i(c))_i$ from Eq.~\eqref{eq:lllite};
$\mathrm{warp}$ is Farneb\"ack flow warping as in
Eq.~\eqref{eq:ew}. \texttt{TE.cached}, \texttt{cond\_emb} are
buffers preserved across batches; refresh schedules are
controlled by \texttt{te\_refresh\_every} ($N_{\text{TE}}$) and
\texttt{cond\_refresh\_every} ($N_{\text{cond}}$).}
\label{alg:loop}
\footnotesize
\begin{tabbing}
xx\=xx\=xx\=xx\=xx\=\kill
\textbf{Input:} batch index $n$, frames $x_{n,1..B}$, prompt $p$, prev. output $y_{n-1}$ \\
\textbf{Side stream} (concurrent with main): \\
\>$z_n \leftarrow \mathrm{VAE.enc}(x_{n,1..B})$ \\
\>\textbf{if} $n \bmod N_{\text{TE}} = 0$ \textbf{or} \texttt{te\_batch\_one} disabled \textbf{then} \\
\>\>$\text{TE.cached} \leftarrow \mathrm{TE}(x_{n,\text{rep}}, p)$ \\
\>\>\>// \texttt{batch\_one}: 1 rep.\ frame, else $B$ \\
\>\textbf{end if} \\
\>$h_n \leftarrow \text{TE.cached}$ \\
\textbf{Main stream}: \\
\>\textbf{if} $n \bmod N_{\text{cond}} = 0$ \textbf{then} \\
\>\>$c_n \leftarrow \mathrm{LLLite.cond}(\mathrm{warp}(y_{n-1}, \mathrm{flow}_{n-1\to n}))$ \\
\>\textbf{else}\ $c_n \leftarrow c_{n-1}$\ \textbf{end if} \\
\>$\hat z_n \leftarrow \mathrm{U\text{-}Net}_{K{=}1}(z_n;\, h_n,\, c_n)$ \\
\>$y_n \leftarrow \mathrm{VAE.dec}(\hat z_n)$ \\
\textbf{Synchronise}: event between side and main streams; \\
\>$\mathrm{wall}_n = \max(t_{\text{side}}, t_{\text{main}})$
\end{tabbing}
\end{algorithm}

\noindent
Cumulatively, the base four speed mechanisms ($a$--$d$;
mode~$e$ is off by default and not used for any reported number
unless explicitly stated) take a no-LLLite baseline from
$1.45$\,\fps{} ($\approx$$690$\,ms/frame) to $32.4$\,\fps{}
(Table~\ref{tab:main}).

\subsection{Compile-Friendly LLLite Reformulation}
\label{sec:method:lllite}

\emph{Principle.} \texttt{torch.compile}~\cite{ansel2024pytorch}
caches a single traced graph and re-uses it only while the forward
pass is \emph{graph-stable}: free of data-dependent Python control
flow and free of changes to the identity of the tensors it closes
over. Temporal-consistency adapters such as
LLLite~\cite{kohya2023lllite} were written for eager execution and
violate both conditions, which is the concrete reason the SD-class
adapter ecosystem does not compose with the compiled fast-denoiser
regime (\S\ref{sec:intro}). The released LLLite forward carries five
data-dependent branches (a \texttt{multiplier=0} short-circuit, a
null-conditioning guard, dtype/device checks, a sequence-length
fallthrough, and a train-only dropout path), and its cached
conditioning embedding is \emph{reassigned} rather than updated
in place on every frame, changing tensor identity and forcing a
Dynamo recompile per frame. The adapter is therefore not slow
because of its arithmetic — it is slow because it defeats graph
caching.

\emph{Reformulation.} We restore graph stability by making the
inference forward (a) branch-free and (b) identity-stable, without
touching the trained weights. Branches are removed by exposing a
dedicated inference path in which the residual $\delta_i$ of
Eq.~\eqref{eq:lllite} is added unconditionally (the pruned dropout,
dtype, and shape-fallback cases never arise at fixed inference
shapes). Identity is stabilised by holding the conditioning
embedding and the multiplier in pre-allocated fixed-shape buffers
that are written in place, so updating the conditioning image no
longer invalidates the graph; existing checkpoints load unchanged.
Implementation-level specifics (buffer shapes, the in-place
\texttt{set\_cond\_image} write, and the
\texttt{forward\_inference} patch mechanism) are given in
Appendix~\ref{app:lllite}. After this refactor an LLLite-augmented
U-Net at $B{=}8$, $K{=}1$ averages $10.12\pm0.11$\,\fps{} across
10 DAVIS sequences versus $2.13\pm0.03$\,\fps{} for the eager LLLite
reference — a $4.75\times$ speed-up with identical outputs.

\subsection{Periodic Conditioning Refresh and Hook Subset}
\label{sec:method:cond}

Even compile-friendly, the 108 LLLite hooks each invoke a small
\textsc{conditioning1} CNN every time \texttt{set\_cond\_image} is
called. At $B{=}8$ this CNN forward is $4\times$ a single image's
cost and dominates the LLLite path's wall time.

\paragraph{Cond-refresh interval} Because consecutive frames share
most scene content, the warped previous output — the LLLite cond
input — changes slowly. We rebuild \texttt{cond\_emb} every $N$-th
batch and reuse the previous embedding in between. With $B{=}8$ and
$N{=}8$ a refresh fires once per 64 frames ($\sim$$2$\,s at
$30$\,\fps{}); the cached embedding remains a good approximation for
intermediate batches and empirically does \emph{not} degrade
reference fidelity or warping error versus per-batch refresh
(\S\ref{sec:exp:quality}). The optical-flow computation that produces
this warped input is moved off the critical path so that it overlaps
the next batch rather than stalling it, and therefore adds no
wall-clock cost (implementation in Appendix~\ref{app:lllite}).

\paragraph{Hook subset} Reducing the active hook count is the second
axis. The 108-hook attachment splits into 38 on
\texttt{down\_blocks}, 16 on \texttt{mid\_block}, and 54 on
\texttt{up\_blocks} on the \dreamlite{} U-Net. Restricting to
\texttt{down\_blocks}-only (38 hooks, $\approx$$35\%$ of the full
set) cuts conditioning-encoder work by the same factor; trained
weights for dropped hooks are silently ignored at load time. The
restriction recovers stylisation sharpness from
Sobel $1.88$ (all-108) to $4.74$ (close to the no-LLLite
baseline $5.10$) and LPIPS-to-teacher from $0.54$ to $0.38$
(Table~\ref{tab:main}). We adopt it as the default operating point.

\paragraph{Caveat on $\varepsilon_w$ in the smoothing regime}
A LLLite configuration with \emph{too much} attribute authority — e.g.,
108 hooks at multiplier $1.0$ — pushes outputs toward a desaturated
``smooth purple'' attractor. The Sobel / HF-FFT / LPIPS triplet in
Table~\ref{tab:main} catches this immediately (108-hook rows:
Sobel $1.88$--$1.97$, HF-FFT $987$--$1046$, LPIPS-to-teacher
$0.52$--$0.54$), while $\varepsilon_w$ \emph{rewards} the smoothing
($12.96$--$13.00$ vs.\ champion's $18.34$) because the output is
nearly static. We therefore recommend always pairing warping error
with at least one spatial high-frequency probe in this regime; the
LCM-LoRA distillation case study reported in \S\ref{sec:neg} is the
direct application of this rule.

\section{Experiments}\label{sec:exp}

\subsection{Setup and Metrics}\label{sec:exp:setup}
Unless noted, all measurements use a single RTX~3090~Ti at
$512\!\times\!512$, $B{=}8$, $K{=}1$, with the temporal LLLite v3
adapter (trained on 10 DAVIS-2017~\cite{pont2017davis} sequences
$\times$50 frames at
the oil-painting prompt for 12 epochs, AdamW8bit, post-hoc
$\alpha{=}0.85$ blended teacher, $\sim$25\,min). We exclude the
first batch of every run (it incurs the \texttt{torch.compile} cost)
and drop partial tail batches (the compiled U-Net and LLLite
\texttt{cond\_emb} buffer are shape-locked to $B$). Throughput
\fps{} is per-frame steady-state wall-clock; warping error
$\varepsilon_w$~\cite{lai2018temporal} is computed with Farneb\"ack
flow on the inputs (we cross-check with RAFT-Large
\cite{teed2020raft} below). We additionally report Sobel mean-abs,
HF-FFT energy outside the centre $H/4$ disc, LPIPS to the eager
4-step teacher~\cite{zhang2018lpips}, and CLIP-sim (the cosine
similarity between each output frame's CLIP image embedding and the
edit prompt's CLIP text embedding, averaged over frames, using the
CLIP ViT-L/14 backbone)~\cite{radford2021clip}. Concurrent edit
benchmarks such as
OpenVE-Bench~\cite{he2025openve3m} have begun to standardise
instruction-guided video editing along Instruction Compliance,
Consistency \& Detail, and Visual Quality \& Stability axes; we
report DAVIS-2017 plus cross-dataset clips here (consistent with the
streaming v2v literature) and note OpenVE-Bench as a natural
cross-comparison target once a matching consumer-GPU baseline lands
on a public leaderboard.

\subsection{Sustained Throughput and Latency}\label{sec:exp:throughput}

Table~\ref{tab:main} reports speed and quality across the same 10
DAVIS-2017 sequences the v3 LLLite was trained on; \emph{all
quality numbers in this table are therefore in-domain} and should
be read as such, with quality generalisation outside the training
set evaluated separately in \S\ref{sec:exp:transfer} (held-out
DAVIS-19, held-out prompts, cross-dataset). The eager 4-step
reference runs at $1.45$\,\fps{}; the full pipeline with all four
speed mechanisms, the trained LLLite, the \texttt{down\_blocks}
hook subset, and cond-refresh runs at
$32.66\!\pm\!1.00$\,\fps{} ($N{=}9$, \textit{scooter-black}
excluded because its 43-frame length is shorter than the $B{=}16$
warm-up$+$measure window) — a $22.5\times$ speed-up. Two
complementary comparisons isolate the adapter contribution: at
identical $K{=}1$, $B{=}16$ the LLLite champion matches the
no-LLLite pipelined baseline in throughput ($-1.2\%$) while cutting
flicker by $20\%$ ($\varepsilon_w$: $22.97\!\to\!18.34$); against a
no-LLLite $K{=}2$ post-hoc flow-warp smoother (the configuration
whose targets we distill for LCM-LoRA training in
\S\ref{sec:neg}), the champion is simultaneously $2.2\times$
faster and $5\%$ lower in $\varepsilon_w$.

\begin{table*}[t]
\caption{Main results on DAVIS-2017 (mean $\pm$ std).
\textbf{fps values are short-window per-batch peaks} on a
3090~Ti (steady-state on 30 measured batches after 3 warm-up
batches); see Table~\ref{tab:sustained} for the
480-frame sustained throughput, and Table~\ref{tab:gpu-scaling}
for the per-GPU hardware-scaling re-measurement.
$^\dagger$ includes the trained Temporal LLLite adapter. The
all-108-hook rows appear to lower $\varepsilon_w$ but also
collapse Sobel ($-60\%$) and HF-FFT ($-38\%$) and inflate
LPIPS-to-teacher ($+30\%$) — the smoothing artifact discussed at
end of \S\ref{sec:method:cond}. \texttt{down\_blocks} restores
spatial detail while keeping the temporal-consistency benefit
($\varepsilon_w$: $18.34$ vs.\ no-LLLite $22.97$).}
\label{tab:main}
\centering
\footnotesize
\setlength{\tabcolsep}{4pt}
\begin{tabular}{lrr@{ }lr@{ }lrrr}
\toprule
\textbf{Configuration} & $N_{\text{seq}}$ & \multicolumn{2}{c}{\textbf{\fps{}}} & \multicolumn{2}{c}{$\varepsilon_w$} & \textbf{Sobel} & \textbf{HF-FFT} & \textbf{LPIPS} \\
\midrule
\multicolumn{9}{l}{\emph{Speed pipeline (no LLLite)}} \\
Eager baseline (4-step)                              & 10 & 1.45  & $\pm$ 0.01 & 25.83 & $\pm$ 3.67 & 4.96 & 1084 & 0.000 \\
+ compile + 2-step + pipelined + blend               & 10 & 15.08 & $\pm$ 0.19 & 19.32 & $\pm$ 3.15 & 4.33 & 1109 & 0.170 \\
+ batched $B{=}8$ + 1-step + pipelined               & 10 & 32.35 & $\pm$ 0.82 & 24.03 & $\pm$ 4.44 & 5.10 & 1596 & 0.398 \\
+ batched $B{=}16$ + 1-step + pipelined              & 9  & 33.04 & $\pm$ 0.70 & 22.97 & $\pm$ 3.77 & 5.16 & 1663 & 0.401 \\
\midrule
\multicolumn{9}{l}{\emph{LLLite-augmented (all-108 hooks; smoothing artifact)}} \\
LLLite v3 eager (108 hooks)$^\dagger$                & 10 & 2.13  & $\pm$ 0.03 & 12.96 & $\pm$ 2.46 & \textit{1.97} & \textit{987} & \textit{0.520} \\
+ compile-friendly LLLite + $B{=}8$ (108)$^\dagger$  & 10 & 10.12 & $\pm$ 0.11 & 13.00 & $\pm$ 2.61 & \textit{1.88} & \textit{1046} & \textit{0.541} \\
\midrule
\multicolumn{9}{l}{\emph{LLLite + \texttt{down\_blocks} hook subset (champion)}} \\
+ cond-refresh + \texttt{down\_blocks}$^\dagger$     & 10 & 30.92 & $\pm$ 0.74 & 19.01 & $\pm$ 3.34 & 4.90 & 1784 & 0.377 \\
\textbf{+ batched $B{=}16$ (champion)}$^\dagger$     & 9  & \textbf{32.66} & $\boldsymbol{\pm}$ \textbf{1.00} & \textbf{18.34} & $\pm$ 2.66 & \textbf{4.74} & \textbf{1760} & \textbf{0.380} \\
\midrule
\multicolumn{9}{l}{\emph{Transfer of the champion configuration to held-out video (see Table~\ref{tab:transfer} for full breakdown)}} \\
\quad unused DAVIS-19 (v3 never trained on)$^\dagger$ & 19 & 29.97 & $\pm$ 0.45 & 19.14 & $\pm$ 4.53 & -- & -- & -- \\
\quad cross-dataset (7 sources, 15 clips)$^\dagger$    & 15 & 30.50 & $\pm$ 0.34 & 19.31 & $\pm$ 4.54 & -- & -- & -- \\
\bottomrule
\end{tabular}
\end{table*}

\noindent
\emph{Reading the table.} The in-domain rows establish the speed
budget; the two transfer rows at the bottom — released v3 adapter
on 19 unused DAVIS sequences and 15 non-DAVIS clips from 7
sources — show that fps and $\varepsilon_w$ on unseen video lie
within the in-clip noise of the in-domain champion. We treat
this, not the in-domain DAVIS-10 row, as the headline operational
result.

\textbf{Sustained throughput and end-to-end latency.} Short eval
windows over-state steady-state throughput, so we report a
480-frame \textit{parkour}-loop run in which per-frame recompilation
is ruled out. Without further optimisation the conditioning-refresh
path settles to $26.4$\,\fps{} but with a bimodal latency profile
(p50 $501$\,ms, p95 $1439$\,ms): the tail is entirely the periodic
refresh, and profiling attributes $96\%$ of each spike to the CPU
optical-flow computation on the refresh batch (848 / 882\,ms),
while the conditioning encoder itself costs only $34$\,ms. Moving
that flow off the critical path so it overlaps subsequent batches
removes the tail and raises sustained throughput to $29.6$\,\fps{}
(p50 $513$\,ms, p95 $756$\,ms, $\sigma{=}81$\,ms); the multi-threaded
flow implementation is detailed in Appendix~\ref{app:lllite}.
End-to-end latency at $30$\,\fps{} input is
$(B{-}1)/30 + \text{batch wall}$: $B{=}16$ p50 $1.01$\,s
(throughput-champion), $B{=}8$ p50 $0.51$\,s (low-latency
operating point at $27.4$\,\fps{} sustained, Table~\ref{tab:sustained}).

\begin{table}[t]
\caption{\textbf{Sustained} throughput and end-to-end latency on
RTX~3090~Ti, \textit{parkour} looped to 480 frames; e2e $=
(B{-}1)/30 + \text{batch wall}$. The threaded-flow optimisation
closes most of the gap between the unoptimised path and the
refresh-disabled upper bound.}
\label{tab:sustained}
\centering
\footnotesize
\setlength{\tabcolsep}{3pt}
\begin{tabular}{@{}p{3.6cm}rrrrr@{}}
\toprule
\textbf{Config} & \textbf{fps} & \textbf{p50} & \textbf{p95}
& \textbf{e2e p50} & \textbf{e2e p95} \\
& & (ms) & (ms) & (ms) & (ms) \\
\midrule
$B{=}16$ refresh$=$8 baseline      & 26.4 & 501 & 1439 & 1001 & 1939 \\
$B{=}16$ refresh$=$8 threaded$+$async & \textbf{29.6} & \textbf{513} & \textbf{756} & 1013 & 1256 \\
$B{=}16$ refresh$=\infty$ (upper bound) & 32.2 & 497 & 506 & 997 & 1006 \\
$B{=}8$\phantom{6} refresh$=$8 threaded$+$async  & 27.4 & 272 & 436 & \textbf{506} & \textbf{669} \\
\bottomrule
\end{tabular}
\end{table}

\textbf{In-pipeline component profile.} Table~\ref{tab:profile}
profiles the \emph{unoptimised} cond-refresh path to isolate the
side/main per-component costs; the sustained champion throughput
\emph{after} the threaded$+$async flow optimisation is reported
separately in Table~\ref{tab:sustained}. Per-component CUDA-event
profiling on the side stream and
\texttt{cuda.synchronize}-bracketed walls on the main stream at
$B{=}16$ yield per-batch wall equal to
$\max(\text{side},\text{main})$ within $0.4\%$ — confirming that
prefetch overlap is operating as designed and that the
configuration is TE-bound. The TE prompt encode at
$29.2$\,ms/frame is the single largest contributor; the
LLLite-augmented U-Net adds only $\sim$$1.7$\,ms/frame over the
no-LLLite U-Net once the graph is hot.

\begin{table}[t]
\caption{In-pipeline component profile of the champion configuration
on RTX~3090~Ti, $B{=}16$, $K{=}1$, \texttt{down\_blocks} LLLite,
refresh$=$8. Side and main columns are simultaneous (one overlaps
the other); the wall row matches $\max(\text{side},\text{main})$
within noise. \textbf{This profile uses the unoptimised
cond-refresh path}: the large $\sigma\!\approx\!306$\,ms on the
wall row is bimodality from the periodic CPU Farneb\"ack flow on
the LLLite refresh batch (one in every eight batches; the mean
$628$\,ms therefore matches the $26.4$\,fps baseline row of
Table~\ref{tab:sustained}, not the $29.6$\,fps threaded-flow row).
Threading the per-frame flow across $8$ worker threads and
async-kicking it during \texttt{prefetch\_batch} reduces the
p95 wall from $1439$ to $756$\,ms (Table~\ref{tab:sustained}); we
keep the unoptimised profile here because it cleanly isolates
the side / main per-component contributions.}
\label{tab:profile}
\centering
\footnotesize
\begin{tabular}{lrr}
\toprule
\textbf{Component} & \textbf{ms/batch} & \textbf{ms/frame} \\
\midrule
\multicolumn{3}{l}{\emph{Side stream (async w.r.t.\ main)}} \\
\quad TE encode (\qwen{}, $B{=}16$) & $466.9\pm 11.8$ & 29.2 \\
\quad VAE encode                     & $161.5\pm 11.1$ & 10.1 \\
\quad \emph{side total}              & \emph{628.4}    & \emph{39.3} \\
\midrule
\multicolumn{3}{l}{\emph{Main stream (sequential within batch)}} \\
\quad U-Net (\texttt{down\_blocks} LLLite, 38) & $302.7\pm 9.1$ & 18.9 \\
\quad VAE decode                                & $190.3\pm 6.8$ & 11.9 \\
\quad \emph{main total}                         & \emph{493.0}   & \emph{30.8} \\
\midrule
$\max(\text{side},\text{main})$                  & 628.4 & 39.3 \\
\textbf{measured wall}                            & $\boldsymbol{628.3 \pm 306.0}$ & \textbf{39.3} \\
\bottomrule
\end{tabular}
\end{table}

\textbf{Isolated batch scaling (TE vs.\ U-Net).}
Table~\ref{tab:batch} reports per-component wall in isolation at
$B\in\{1,2,4,8\}$. The TE shows strong sub-linear scaling
($s{=}t_B / (B\!\cdot\!t_1)$ drops from $1.000$ at $B{=}1$ to
$0.233$ at $B{=}8$, i.e.\ $4.3\times$ throughput), while the
compiled U-Net saturates near $s{=}0.77$ at the same batch.
The asymmetry is the direct empirical warrant for buffering
$B$ frames before each pipeline call: the TE benefits
super-linearly while the U-Net cost is near-flat per unit time.

\begin{table}[t]
\caption{Per-component batch scaling in isolation on RTX~3090~Ti.
Sub-linearity $s{=}t_B / (B \!\cdot\! t_1)$; smaller is better
($0.5$ would be a perfect $2\times$ throughput per unit time).
The TE benefits substantially from batching; the compiled U-Net
is already close to peak.}
\label{tab:batch}
\centering
\footnotesize
\begin{tabular}{lrrrr}
\toprule
& $B{=}1$ & $B{=}2$ & $B{=}4$ & $B{=}8$ \\
\midrule
\textbf{TE} (eager, fp16) ms        & 125 & 136 & 153 & 233 \\
\quad sub-linearity $s$              & 1.000 & 0.544 & 0.306 & \textbf{0.233} \\
\textbf{U-Net} (compiled, fp16) ms  & 23  & 41  & 74  & 142 \\
\quad sub-linearity $s$              & 1.000 & 0.891 & 0.804 & 0.772 \\
\bottomrule
\end{tabular}
\end{table}

\textbf{Locating our number against StreamDiffusion (same stack).}
The published StreamDiffusion~\cite{kodaira2023streamdiffusion}
numbers (91\,\fps{} SD-Turbo $K{=}1$ text-to-image; 38\,\fps{}
Kohaku v2.1$+$LCM-LoRA $K{=}4$ image-to-image, both at
$512^2$ on RTX~4090 with TensorRT) come from a different model
class (CLIP-class text-only encoder vs.\ a $2.13$\,B vision-aware
MLLM). For a same-stack reference we re-ran their released benchmark
on an RTX~4090 with their full TensorRT path engaged, porting it to a
current TensorRT toolchain so that the comparison runs on identical
hardware and acceleration (the three compatibility patches required
to build their engines under TensorRT~10.x, and the exact host
configuration, are listed in Appendix~\ref{app:lllite}). With the
TRT engines built, Kohaku v2.1$+$LCM-LoRA reaches a steady-state mean of
$43.6\!\pm\!0.6$\,\fps{} at $K{=}2$ and $29.9\!\pm\!0.3$\,\fps{} at
$K{=}4$ (100 iter, 10 warm-up). On the \emph{same} 4090 host our
champion measures $54.9\!\pm\!0.6$\,\fps{}; on RTX~3090~Ti without
TRT we measure $32.7$\,\fps{} against StreamDiffusion's
$9.3$\,\fps{} (Kohaku $K{=}2$, xformers only). We report these as
same-stack systems context, not as a benchmark superiority claim,
because the model classes and conditioning modalities differ.

\textbf{Head-to-head against an independent V2V system (StreamV2V).}
The StreamDiffusion reference above shares our own stack; to compare
against an \emph{independent} streaming V2V method we ran official
StreamV2V~\cite{liang2025streamv2v} (SD-1.5 $+$ LCM-LoRA with a
cross-frame feature bank) on the \emph{pixel-identical} 512$^2$
inputs our champion consumes, on the same RTX~3090~Ti, matched to the
oil-painting style via the authors' released style LoRA.
Both systems run at the \texttt{xformers} acceleration tier, and this
is not a handicap we impose on the baseline: StreamV2V's feature bank
--- the mechanism responsible for its temporal consistency --- is
\emph{structurally incompatible with TensorRT in the authors' own
implementation}, which raises \texttt{NotImplementedError} when
cached attention and TensorRT are combined. The feature-bank system
therefore runs \emph{only} at this tier; the authors' published
TensorRT figure ($20$\,\fps{}, A100) is the feature-bank-\emph{off}
configuration, which by their own documentation reduces to per-frame
StreamDiffusion (the same model class as our same-stack reference
above). Our own regime independently cannot use TensorRT
(\S\ref{sec:intro}, MLLM-TE FP16 overflow). We therefore read
Table~\ref{tab:streamv2v} as a \emph{throughput frontier per GPU
class}, not a quality contest, and flag two confounds explicitly
below.

\begin{table}[t]
\centering
\caption{Independent-baseline comparison on 10 DAVIS-2017 clips,
$512{\times}512$, RTX~3090~Ti, feature bank on for StreamV2V (its
TensorRT path is incompatible with the bank; \S\ref{sec:exp:throughput}),
mean~$\pm$~s.d. Warping error $\varepsilon_w$ is reported \emph{with}
a frequency-domain panel and is not read alone. Sobel edge energy is
identical between the two systems, so neither is in the
smoothing-collapse regime of \S\ref{sec:exp:quality}; the
$\varepsilon_w$ gap instead tracks a texture difference --- our output
adds high-frequency stylization detail relative to the source
($+0.16$ log-ratio) while StreamV2V slightly reduces it ($-0.05$),
and $\varepsilon_w$ penalises added HF detail. We therefore do
\emph{not} claim superior temporal quality; StreamV2V may be
genuinely smoother, and we report the trade-off rather than adjudicate
it. Sobel/HF-FFT here use per-frame $[0,1]$-normalised intensities and
a source-referenced log-ratio, a different scale from the absolute
Sobel/HF values in Table~\ref{tab:main} (within-panel comparison
only).}
\label{tab:streamv2v}
\small
\begin{tabular}{lccc}
\toprule
System & fps~$\uparrow$ & $\varepsilon_w$ & HF-FFT vs.\ src \\
\midrule
StreamV2V~\cite{liang2025streamv2v} & $4.4\!\pm\!0.3$ & $14.0\!\pm\!5.2$ & $-0.05$ \\
\dreamlite{}-stream (ours) & $\mathbf{29.7\!\pm\!1.5}$ & $19.8\!\pm\!3.4$ & $+0.16$ \\
\midrule
DAVIS source (Sobel ref.) & --- & --- & $0.00$ \\
\bottomrule
\end{tabular}
\end{table}

\noindent
At the same acceleration tier our pipeline sustains $6.8\times$ the
throughput of StreamV2V on identical inputs, while both systems match
in coarse edge energy (Sobel mean-abs $0.169$ for both vs.\ $0.239$
for the source). \emph{Two confounds temper this number.} First, part
of the gap is architectural rather than attributable to our
pipelining: we batch $B{=}16$, whereas StreamV2V's causal cross-frame
bank is inherently sequential and cannot batch, so the comparison
measures our batched-throughput advantage as well as the pipelining
contribution. Second, on the one axis where StreamV2V leads
($\varepsilon_w$) we concede it may produce genuinely smoother output;
we could push our own $\varepsilon_w$ toward StreamV2V's by
cross-frame noise alignment (a fixed-noise schedule that cut temporal
drift markedly in a companion probe) but this trades stylization
sharpness for smoothness, and we report the champion's operating point
rather than the $\varepsilon_w$-minimising one. For reproducibility we
note that the current StreamV2V \texttt{main} pins a rebased
\texttt{diffusers} fork mutually unsatisfiable with its own
\texttt{torch==2.1} constraint; the ICLR-2025 commit
(\texttt{diffusers==0.27.0}, attention-processor feature bank) runs as
released, and our harness is provided in the supplement.
We do not run Live2Diff~\cite{xing2024live2diff} or
StreamDiffusionV2~\cite{feng2025streamdiffusionv2} head-to-head: both
are video-diffusion (DiT) systems in a different backbone class from
our distilled image-diffusion edit U-Net, and StreamDiffusionV2 in
particular assumes a multi-GPU rolling-KV-cache deployment, so a
single-consumer-GPU rerun would misrepresent them; we position against
them qualitatively in Table~\ref{tab:streamv2v-landscape} instead.
Finally, the identical Sobel yet divergent HF-FFT between the two
systems is not a contradiction: Sobel mean-abs is a broadband
gradient measure dominated by mid frequencies, whereas the HF-FFT
log-ratio isolates energy \emph{outside} the centre $H/4$ disc, so two
outputs can match in coarse edge energy while differing in the highest
band.

\subsection{Temporal Quality, Flow Estimator, and Scene Cuts}\label{sec:exp:quality}

\textbf{Cond-refresh sweep.} Table~\ref{tab:refresh} sweeps
$N\in\{1,4,8,16\}$ on the champion \texttt{down\_blocks}
configuration ($B{=}8$, $K{=}1$, 38 hooks). Throughput climbs from
$25.4$ to $28.9$\,\fps{} while $\varepsilon_w$, Sobel and HF-FFT
are flat to three significant figures and LPIPS-to-$N{=}1$ rises
only from $0$ to $0.030$ at $N{=}16$ — small compared with the
LPIPS-to-4-step-teacher of $\sim$$0.40$ at the same operating
point. The same monotone-then-saturated trend holds for the
108-hook configuration (gap from $N{=}1$ to $N{=}8$ is larger
because the per-hook \texttt{conditioning1} CNN dominates the
refresh batch; see supplementary). We adopt $N{=}8$.

\begin{table}[t]
\caption{Cond-refresh sweep on the champion \texttt{down\_blocks}
configuration ($B{=}8$, $K{=}1$, 38 hooks). $\varepsilon_w$, Sobel
and HF-FFT are flat across $N$; LPIPS-to-$N{=}1$ at $N{=}16$ is
$\sim$$0.03$, vs.\ LPIPS-to-4-step-teacher of $\sim$$0.40$ at the
same operating point.}
\label{tab:refresh}
\centering
\footnotesize
\setlength{\tabcolsep}{3pt}
\resizebox{\columnwidth}{!}{%
\begin{tabular}{rrrrrr}
\toprule
$N$ & \textbf{fps} & $\boldsymbol{\varepsilon_w}$
& \textbf{Sobel} & \textbf{HF-FFT} & \textbf{LPIPS$_{N{=}1}$} \\
\midrule
1  & $25.36 \pm 0.25$ & $19.26 \pm 3.61$ & 4.86 & 1725 & 0.000 \\
4  & $27.67 \pm 1.14$ & $19.25 \pm 3.61$ & 4.87 & 1727 & $0.025 \pm 0.004$ \\
\textbf{8} & $\boldsymbol{28.21 \pm 1.59}$ & $19.26 \pm 3.60$ & 4.86 & 1727 & $0.027 \pm 0.006$ \\
16 & $28.90 \pm 1.50$ & $19.26 \pm 3.64$ & 4.87 & 1731 & $0.030 \pm 0.008$ \\
\bottomrule
\end{tabular}%
}
\end{table}

\textbf{Long-sequence drift.} DAVIS clips are short (43--100
frames); to verify stability over a longer streaming window
than any single DAVIS clip permits, we loop \textit{parkour} to
480 frames ($20$\,s at $24$\,\fps{}) and chunk the measured 448
frames into seven 64-frame windows. A truly multi-minute
sustained test remains future work. Table~\ref{tab:drift} reports
$\varepsilon_w$, Sobel, HF-FFT, and LPIPS-against-chunk-0; the
seven-chunk std on each metric stays within the inter-clip noise
of Table~\ref{tab:main} (no monotonic drift). LPIPS shows a
one-time jump at chunk 1 (different per-frame noise produces
different per-frame stylisation even for repeating inputs) but is
stable across chunks 1--6.

\begin{table}[t]
\caption{Long-sequence drift on \textit{parkour} looped to 480
frames (champion configuration, 2-batch warm-up dropped, then
chunked by 64 frames). Per-chunk std on each metric is within the
inter-clip noise of Table~\ref{tab:main}.}
\label{tab:drift}
\centering
\footnotesize
\begin{tabular}{rrrrrr}
\toprule
\textbf{chunk} & \textbf{frames} & $\boldsymbol{\varepsilon_w}$
& \textbf{Sobel} & \textbf{HF-FFT} & \textbf{LPIPS$\to$ch0} \\
\midrule
0 & 0--64    & 17.62 & 5.30 & 1755 & 0.000 \\
1 & 64--128  & 17.45 & 5.00 & 1736 & 0.724 \\
2 & 128--192 & 17.21 & 4.87 & 1669 & 0.725 \\
3 & 192--256 & 17.65 & 5.37 & 1782 & 0.640 \\
4 & 256--320 & 17.93 & 4.82 & 1690 & 0.728 \\
5 & 320--384 & 18.02 & 5.06 & 1715 & 0.713 \\
6 & 384--448 & 17.38 & 5.34 & 1791 & 0.694 \\
\midrule
\textbf{7-chunk std} & & \textbf{0.27} & \textbf{0.21}
& \textbf{45} & \textbf{0.030$^\dagger$} \\
\bottomrule
\end{tabular}
\\[1pt]
{\scriptsize $^\dagger$ std over chunks 1--6 only (chunk 0 is the
reference).}
\end{table}

\textbf{Robustness to the flow estimator.}
We recompute $\varepsilon_w$ on champion outputs using RAFT-Large
\cite{teed2020raft}
(\texttt{torchvision} \texttt{raft\_large} with
\texttt{C\_T\_SKHT\_V2} weights). On the 9-clip champion subset the two
estimators' per-clip means agree to within
$-0.7\pm 5.6\%$ (RAFT $18.07\pm 2.02$ vs.\ Farneb\"ack
$18.29\pm 2.71$). On the three rows of Table~\ref{tab:main} that
span the no-LLLite / champion / 108-hook smoothing-artifact
spectrum, both estimators rank the configurations identically and
the per-aggregate relative difference is $\le 2\%$, so the ordering
claims are not Farneb\"ack-specific.

\begin{figure}[t]
\centering
\includegraphics[width=\columnwidth]{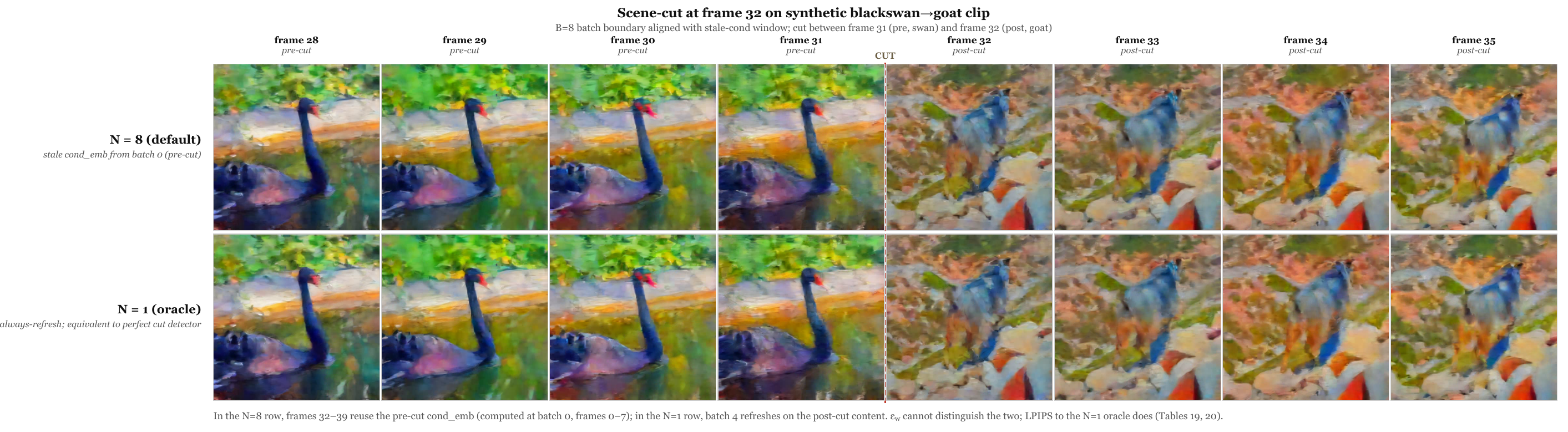}
\caption{Scene-cut at frame~32 on the synthetic
\texttt{blackswan\_goat} clip. \textbf{Top:} $N{=}8$ default
cond-refresh (the stale embedding from batch~0 was computed on
pre-cut swan content). \textbf{Bottom:} $N{=}1$ always-refresh
oracle (refreshes every batch, including batch 4 on post-cut goat
content). The visual divergence between the two rows is small in
the pre-cut window (frames 28--31, both styling the same swan
content) and remains small in the post-cut window
(frames 32--35) — small enough that $\varepsilon_w$ does not
resolve it ($\le 0.11$ absolute, $<\!0.7\%$ relative) but large
enough that LPIPS against $N{=}1$ does ($0.020\!\to\!0.030$,
$\sim$$50\%$ relative jump). The cut is marked with a dashed
vertical line.}
\label{fig:scenecut}
\end{figure}

\textbf{Scene-cut robustness.} Periodic cond-refresh with $N{=}8$
can leave the LLLite embedding up to $N{-}1$ batches stale at a
hard scene change. We constructed three synthetic hard-cut clips
(\texttt{blackswan\_goat}, \texttt{kite\_dance},
\texttt{libby\_camel}; first 32 frames of one DAVIS sequence
concatenated with first 32 of another) timed so the four post-cut
batches all reuse the pre-cut \texttt{cond\_emb} (the worst
possible alignment). On these clips $\varepsilon_w$ cannot
distinguish $N{=}8$ from $N{=}1$ (per-clip absolute
$\le 0.11$, $<\!0.7\%$ relative in every window) because a
stale-but-fixed cond still produces a temporally consistent output.
LPIPS-against-$N{=}1$ does surface the divergence: pre-cut
$0.020\pm 0.002$ vs.\ post-cut 8-frame window $0.030\pm 0.004$
($\sim$$50\%$ relative increase, reproduced on all three clips;
Fig.~\ref{fig:scenecut} shows the visual pattern on
\texttt{blackswan\_goat}). The takeaway is a metric one:
$\varepsilon_w$ alone is insufficient to certify cut-robustness;
a production deployment for cut-heavy streams should trigger an
explicit refresh on cut detection.

\begin{table}[t]
\caption{Scene-cut robustness: $\varepsilon_w$ and
LPIPS-against-oracle on three synthetic hard-cut clips at
worst-case timing (cut at frame 32, mid-stale window).
$N{=}8$ is the default cond-refresh interval; $N{=}1$ is the
always-refresh oracle equivalent to a perfect cut detector.
$\varepsilon_w$ does not distinguish the two ($\le 0.11$ absolute),
but LPIPS against $N{=}1$ surfaces a $\sim$$50\%$ post-cut
relative jump, reproduced on each clip.}
\label{tab:scenecut}
\centering
\footnotesize
\setlength{\tabcolsep}{3pt}
\resizebox{\columnwidth}{!}{%
\begin{tabular}{lrrrr}
\toprule
\textbf{Clip} & $\boldsymbol{\varepsilon_w^{N=8}}$
& $\boldsymbol{\varepsilon_w^{N=1}}$
& \textbf{LPIPS pre} & \textbf{LPIPS post-8} \\
\midrule
blackswan\_goat & 17.89 & 17.78 & 0.0216 & 0.0319 \\
kite\_dance     & 20.71 & 20.80 & 0.0177 & 0.0335 \\
libby\_camel    & 17.33 & 17.30 & 0.0220 & 0.0259 \\
\midrule
mean $\pm$ std  & $18.64\pm 1.81$ & $18.63\pm 1.89$
                & $0.020\pm 0.002$ & $\boldsymbol{0.030\pm 0.004}$ \\
\bottomrule
\end{tabular}%
}
\end{table}

\subsection{Held-Out and Cross-Dataset Transfer}\label{sec:exp:transfer}

\begin{figure}[t]
\centering
\includegraphics[width=\columnwidth]{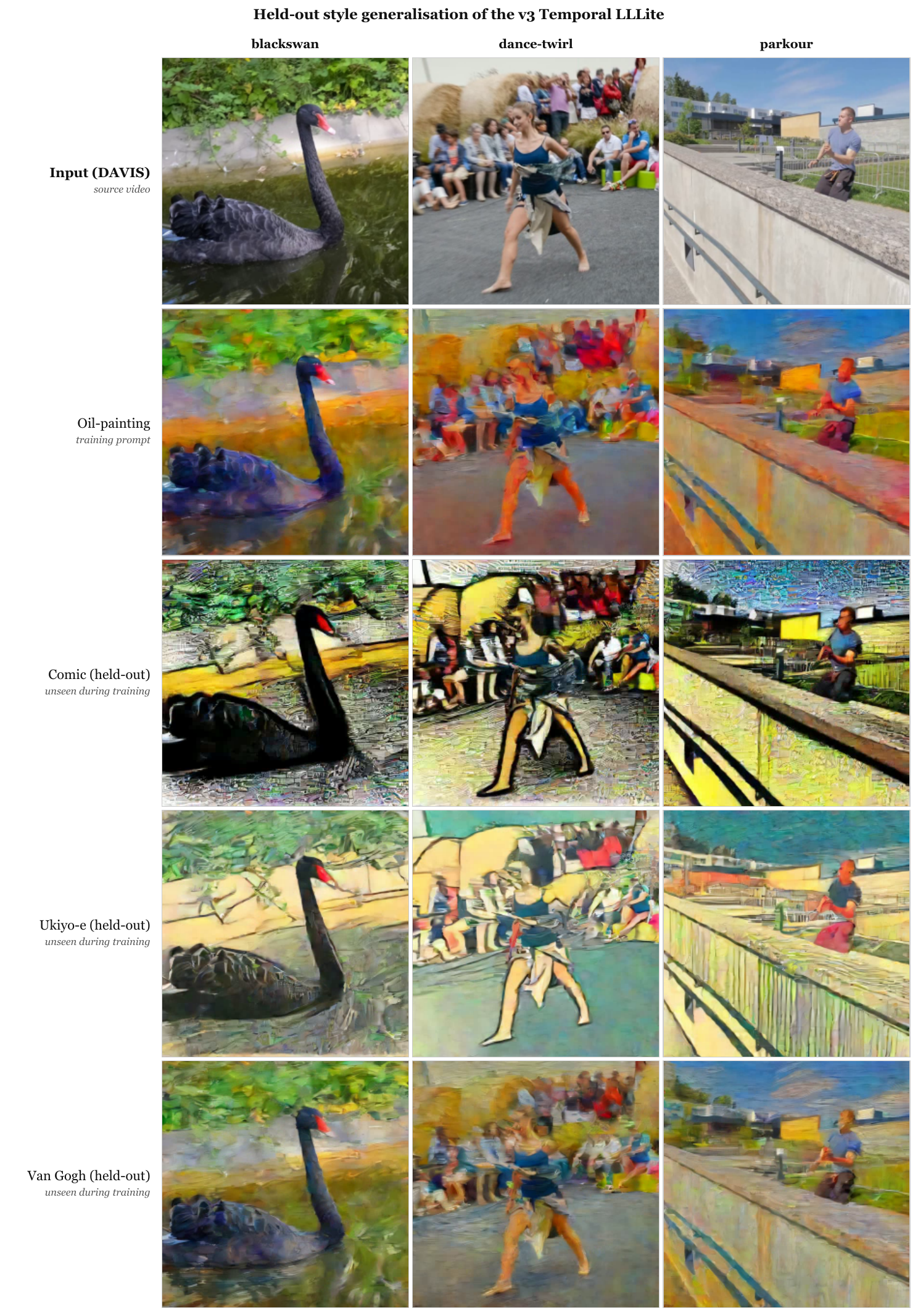}
\caption{Qualitative held-out-prompt results on three DAVIS
clips (\textit{blackswan}, \textit{dance-twirl},
\textit{parkour}). Top: input; second row: in-domain oil-painting
(training prompt); bottom three rows: held-out prompts on the same
v3 adapter, never seen during training. The visual pattern tracks
Table~\ref{tab:transfer}: \textit{van Gogh} generalises cleanly,
\textit{ukiyo-e} captures flat-colour line-art partially, and
\textit{comic halftone} \textbf{is the principal failure mode of
the v3 adapter} — the halftone-dot texture is unlike any
training-prompt style, the LLLite cond-encoder was not exposed to
it, and the output collapses to a noisy high-frequency rendering
(HF-FFT $9189$ vs.\ in-domain $1760$). Single mid-clip frame
($t{=}16$) per cell.}
\label{fig:qualitative}
\end{figure}

Table~\ref{tab:transfer} consolidates four transfer experiments
along three axes: held-out videos within DAVIS, unused DAVIS
sequences for the released v3 adapter, held-out prompts, and
non-DAVIS sources. (i) A separate \emph{v5-heldout} adapter trained
on 7 of 10 DAVIS sequences and evaluated on the held-out 3 reaches
$\varepsilon_w$ $19.14\pm 0.68$ versus v3's in-domain $21.05\pm 2.04$
on the same 3 sequences — within in-clip noise. (ii) Evaluating the
released v3 adapter on 19 additional DAVIS sequences it never saw
during training yields fps $29.97\pm 0.45$, $\varepsilon_w$
$19.14\pm 4.53$, within inter-clip noise of the in-domain
DAVIS-10 champion ($30.07\pm 0.37$, $18.96\pm 3.34$). (iii) On 15
clips from 7 non-DAVIS sources (Big~Buck~Bunny CC-BY $\times$3,
Sintel CC-BY $\times$1, Jellyfish $\times$3, MDN CC0 $\times$2,
learning-container $\times$1, Intel IoT sample-videos $\times$5)
fps reaches $30.50\pm 0.34$ and $\varepsilon_w$ $19.31\pm 4.54$,
again indistinguishable from in-domain DAVIS. (iv) On three
held-out style prompts the gradient tracks semantic distance to the
training prompts — \textit{van Gogh} matches the in-domain champion
on every metric, \textit{ukiyo-e} degrades roughly $2\times$, and
\textit{comic halftone} fails (high-frequency texture outside the
adapter's training manifold). CLIP-sim
\cite{radford2021clip} provides an
independent prompt-adherence signal that ranks the four prompts in
the same order (Table~\ref{tab:transfer}, last column).
Figure~\ref{fig:qualitative} shows the visual pattern on three
DAVIS clips.

\begin{table}[t]
\caption{Transfer evaluation. (a) v5-heldout: separate adapter
trained on 7 DAVIS, evaluated on 3 held-out. (b) v3 on 19
additional unused DAVIS sequences. (c) v3 on 15 non-DAVIS clips
(7 sources). (d) the released v3 adapter on 3 held-out
style prompts it never saw in training (prompt generalisation; the
separately-trained multi-prompt v4 adapter is analysed on its own in
Table~\ref{tab:multiprompt}). All use the champion configuration. The
in-domain $\varepsilon_w$ is $18.96$ on DAVIS-10 here vs.\ $18.34$ in
Table~\ref{tab:main}, which reports DAVIS-9 (excluding the short
43-frame \textit{scooter-black} clip that is shorter than the
$B{=}16$ warm-up$+$measure window).}
\label{tab:transfer}
\centering
\footnotesize
\setlength{\tabcolsep}{3pt}
\begin{tabular}{@{}llrrrr@{}}
\toprule
\textbf{Axis} & \textbf{Set} & $N_{\text{seq}}$ & \textbf{fps}
& $\boldsymbol{\varepsilon_w}$ & \textbf{CLIP} \\
\midrule
in-domain & DAVIS-10 champ.  & 10 & 30.07 & 18.96 & 0.218 \\
\midrule
held-out video & v5h, 3 clips    &  3 & 33.06 & 19.14 & --    \\
unused DAVIS   & v3, 19 clips    & 19 & 29.97 & 19.14 & --    \\
cross-dataset  & 7 sources, 15 clips & 15 & 30.50 & 19.31 & --    \\
\midrule
held-out prompt& \textit{van Gogh} &  9 & 30.15 & 18.75 & 0.232 \\
held-out prompt& \textit{ukiyo-e}  &  9 & 30.40 & 37.43 & 0.156 \\
held-out prompt& \textit{comic halftone} & 9 & 30.36 & 51.02 & 0.172 \\
\bottomrule
\end{tabular}
\end{table}

\paragraph{Multi-prompt v4 LLLite per-prompt breakdown}
The held-out-prompt rows in Table~\ref{tab:transfer}(d) test the
\emph{released v3} adapter (trained on oil painting only) on prompts
it never saw. Separately, to probe multi-style capacity, we trained a
distinct adapter on five style prompts simultaneously
(oil painting, watercolor, pencil sketch, anime cel-shaded,
3D-render ray-traced) — 110 distillation pairs per prompt,
$550$ pairs total, otherwise identical to the v3 recipe. We refer
to this adapter as \emph{v4}; it is analysed only in this paragraph
and Table~\ref{tab:multiprompt}, and is not used anywhere else in
the paper. Table~\ref{tab:multiprompt} reports
the v4 per-prompt breakdown on the same DAVIS-9 split as the
champion. v4 maintains v3's in-domain quality on oil painting
($\varepsilon_w$ $18.21$ vs.\ v3's $18.34$) and reaches
$30+$\,\fps{} on every prompt, with $\varepsilon_w$ tracking the
natural high-frequency content of each style. The 3D-render
prompt is slightly slower because the encoding for an MLLM
instruction containing two highlighted concepts (3D + ray tracing)
sits at the high end of the TE compute distribution; the U-Net
path is unchanged.

\begin{table}[t]
\caption{Multi-prompt v4 LLLite per-prompt breakdown on DAVIS-9
($N_{\text{seq}}{=}9$). v4 holds in-domain quality on oil painting
($\varepsilon_w$ $18.21$ vs.\ v3's $18.34$) and runs at
$30+$\,\fps{} on every prompt; per-prompt $\varepsilon_w$ tracks
each style's natural high-frequency content (sharp pencil lines
flicker more than soft watercolor edges).}
\label{tab:multiprompt}
\centering
\footnotesize
\setlength{\tabcolsep}{4pt}
\begin{tabular}{lrr}
\toprule
\textbf{Prompt} & \textbf{fps} & $\boldsymbol{\varepsilon_w}$ \\
\midrule
v3 single (oil)$^\dagger$            & $32.66 \pm 1.00$ & $18.34 \pm 2.66$ \\
\midrule
oil painting$^\dagger$               & $32.25 \pm 0.41$ & $18.21 \pm 1.92$ \\
watercolor$^\dagger$                 & $32.44 \pm 0.30$ & $15.46 \pm 3.90$ \\
pencil sketch$^\dagger$              & $32.41 \pm 0.41$ & $28.08 \pm 3.64$ \\
anime cel-shaded$^\dagger$           & $32.16 \pm 0.35$ & $24.77 \pm 5.18$ \\
3D-render ray-traced$^\dagger$       & $29.72 \pm 0.25$ & $23.43 \pm 4.29$ \\
\midrule
\textbf{v4 overall} (5 prompts $\times$ 9 seq.)$^\dagger$
                                     & $\boldsymbol{31.80 \pm 1.11}$
                                     & $21.99 \pm 5.94$ \\
\bottomrule
\end{tabular}
\\[1pt]
{\scriptsize $^\dagger$ champion configuration ($B{=}16$, $K{=}1$,
\texttt{down\_blocks}, refresh $N{=}8$).}
\end{table}

\subsection{Hardware Scaling}\label{sec:exp:hw}

Table~\ref{tab:gpu-scaling} reports the champion configuration on
three consumer NVIDIA GPUs at the same operating point, same v3
LLLite checkpoint, same 10 DAVIS sequences. RTX~4090 (Ada Lovelace)
has the \emph{same} $1008$\,GB/s memory bandwidth as the 3090~Ti yet
reaches $54.9\pm 0.6$\,\fps{} ($+83\%$); because bandwidth is
unchanged, this gain is attributable to Ada's higher clocks, larger
L2 cache, and roughly doubled FP16 throughput. RTX~5090 (Blackwell,
$+78\%$ memory bandwidth, $+34\%$ TFLOPs, torch nightly $2.12$-cu128
for SM\_120 support) reaches $\textbf{74.1}\!\pm\!4.5$\,\fps{}
($+147\%$ over 3090~Ti, $+35\%$ over 4090). Both cross-generation
gains track compute (clocks, cache, FP16 throughput) rather than
memory bandwidth. This refines the profile of \S\ref{sec:exp:throughput}:
the per-batch wall is dominated by the TE \emph{component}
(Table~\ref{tab:profile}), but that component is compute- and
kernel-bound, not memory-bandwidth-bound --- so throughput scales
with a GPU's compute and cache, not its bandwidth.

\begin{table}[t]
\caption{\textbf{Hardware scaling} of the champion configuration
($B{=}8$, $K{=}1$, \texttt{down\_blocks}$+$refresh, BF16). fps
values are short-window steady-state across the same 10 DAVIS-10
sequences as Table~\ref{tab:main}; not a 480-frame sustained
measurement. Same v3 LLLite checkpoint; the RTX~5090
row additionally uses the PyTorch-nightly stack disclosed in
\S\ref{sec:exp:repro}.}
\label{tab:gpu-scaling}
\centering
\footnotesize
\setlength{\tabcolsep}{3pt}
\resizebox{\columnwidth}{!}{%
\begin{tabular}{lcccc}
\toprule
\textbf{GPU} & \textbf{Gen.} & \textbf{Mem.\ BW} & \textbf{\fps{}} & \textbf{$\varepsilon_w$} \\
\midrule
RTX~3090~Ti & Ampere    & 1008\,GB/s & $30.07\pm 0.37$ & $18.96\pm 3.34$ \\
RTX~4090    & Ada       & 1008\,GB/s & $54.9\pm 0.6$   & $18.96\pm 3.34$ \\
\textbf{RTX~5090} & \textbf{Blackwell} & \textbf{1792\,GB/s} & $\boldsymbol{74.1\pm 4.5}$ & $19.73\pm 3.47$ \\
\bottomrule
\end{tabular}%
}
\end{table}

\subsection{Reproducibility}\label{sec:exp:repro}
All measurements were produced from a single commit of the public
inference repository (\texttt{otanl/dreamlite-stream}, commit hash
\texttt{efe2caa}); the code state is archived as a Zenodo release at
\href{https://doi.org/10.5281/zenodo.20389428}{DOI 10.5281/zenodo.20389428},
and the released v3 LLLite adapter weights are attached to the
GitHub release with the same tag.
The 3090~Ti and 4090 runs use Windows~11~Pro,
PyTorch~2.6.0+cu124, Triton~3.2 (Windows port), and CUDA driver
$\geq$$565$. The RTX~5090 row additionally requires PyTorch
nightly $\geq$$2.12$-cu128 for SM\_120 compute-capability support;
the exact wheel pinning is recorded in
\texttt{environment\_5090.yml} alongside the repository, and we
disclose this explicitly because the nightly stack is unstable
across snapshots. The v3 LLLite adapter checkpoint
(\texttt{lllite\_v3.safetensors}) and its SHA-256 are recorded in
the supplementary \texttt{config.json} together with the prompt
list and DAVIS sequence split. Each table number is backed by a
per-sequence JSONL file in the supplementary; the included
\texttt{scripts/champion\_eval.py},
\texttt{scripts/sustained\_throughput\_test.py}, and
\texttt{scripts/cross\_dataset\_eval.py} regenerate
Tables~\ref{tab:main}, \ref{tab:sustained}, \ref{tab:profile},
\ref{tab:batch}, \ref{tab:refresh}, \ref{tab:drift},
\ref{tab:scenecut}, \ref{tab:transfer}, \ref{tab:multiprompt},
\ref{tab:gpu-scaling}, and \ref{tab:lcmlora} from raw DAVIS and
non-DAVIS inputs. Warm-up rules (first batch excluded; partial
tail batches dropped) are identical across all reported numbers.

\textbf{Base-model availability.} The pipeline depends on the
upstream \dreamlite{} base-model checkpoint
\cite{feng2026dreamlite}. At submission time the inference code
is released under an open licence, while the base-model weights
are listed on a release plan with the weights themselves gated
by a safety-review / request process; we used the bf16/fp16
weights available to us at the time of these experiments and
record the specific checkpoint hash in our supplementary
\texttt{config.json}. We do not claim public reproducibility of
the upstream base-model weights themselves; we release the
inference code, evaluation harness, our v3 Temporal-LLLite
adapter weights, the LLLite-baked TRT export tooling, and the
benchmark / ablation scripts. Readers should consult the
\dreamlite{} and \qwen{} release notes for current access
modality and licence constraints, as these may change after
submission.

\section{Negative Results and Case Studies}\label{sec:neg}

We summarise five engineering dead ends and one distillation case
study; each ruled out a tempting direction and shaped the final
recipe.

\paragraph{Naive previous-latent init mixing}
Mixing the previous output latent into the noisy init at frame $t$
(a standard SD streaming trick~\cite{kodaira2023streamdiffusion})
degenerates within $\sim$10 frames into a self-reinforcing single-colour
attractor on the 4-step distilled flow-matching scheduler. The
4-step schedule lacks the integration steps to pull the trajectory
back on-manifold once the init is off-distribution. $\varepsilon_w$
falsely drops (the output is near-static), but reference fidelity
catches the failure: $L_1^{\text{ref}}$, the mean per-pixel $L_1$
between the streamed output and an independent per-frame full-quality
denoise of the same input, cannot be gamed by a near-static output
(unlike $\varepsilon_w$) and rises sharply here.

\paragraph{Cond-image manipulation in distilled edit models}
Replacing the right half of the spatial-concat \textsc{edit} input
with a flow-warped previous output, a learned T2I keyframe, or a
blend all break: the distilled U-Net trained with the cond-image
matching the input distribution drifts to an abstract red/orange
blur when fed a stylised intermediate or a hallucinated keyframe.
Cond-side conditioning is not a free temporal channel.

\paragraph{Pure flow-warped speculation}
Speculatively predicting frame $t{+}1$ by warping frame $t$'s latent
and skipping the U-Net unless flow magnitude crosses a threshold,
in the spirit of speculative decoding~\cite{leviathan2023speculative},
reaches the noise floor of the base flicker without a
temporally-consistent base ($\sim$30 vs.\ input flicker $\sim$28).
$\varepsilon_w$ is constructively low ($\sim$11) because the output
is by construction a flow-warp of the previous frame; the metric is
invariant under exactly that warp. We use this as motivation for
cond-refresh: amortising the LLLite cost via $N{>}1$ is much cheaper
than a learned skip mechanism.

\paragraph{TensorRT FP16 with wide-range MLLM TE outputs}
TensorRT does not give us a faster path than PyTorch$+$compile here,
for a reason specific to vision-aware MLLM conditioning: the text
encoder emits activations with an unusually wide dynamic range
(std $37$, spanning $-854$ to $1160$), and once these are projected
into the U-Net attention dimension they overflow the half-precision
attention kernels that TensorRT selects, producing visibly broken
outputs. The standard precision-pinning remedies do not recover
accuracy. Forcing the whole engine to full precision is numerically
equivalent to PyTorch (cos-sim $0.999996$) but runs at $23.7$\,\fps{},
slower than PyTorch$+$compile. Exposing the adapter conditioning
directly in the engine is structurally sound (FP32 cos-sim
$0.999999$; FP16 at $2.14\times$ the PyTorch U-Net wall) yet hits the
same half-precision attention overflow, with fidelity collapsing to
cos-sim $0.77$ on real embeddings (precision-hint list and export
details in Appendix~\ref{app:lllite}). FP8-capable hardware (H100, Blackwell) with explicit
per-tensor attention scaling is a plausible direction we have not
verified.

\paragraph{TE template-token pruning absorbed by side-stream slack}
Removing the 80-token \qwen{} system prompt from the chat template
cuts TE wall by $13\%$ ($262\to229$\,ms at $B{=}8$) in isolation,
but does not move end-to-end fps ($30.09\to30.10$): the saving is
fully absorbed by side-stream slack because the main stream sets
the wall. Quality regresses slightly ($\varepsilon_w$
$18.4\to19.9$), consistent with the system prompt being
load-bearing for instruction-tuned behaviour. \emph{Lesson:} once
side-stream pipelining is in place, the per-frame bottleneck migrates
to the slower stream and improvements to the faster stream are
invisible without also accelerating the slower one.

\begin{figure}[t]
\centering
\includegraphics[width=\columnwidth]{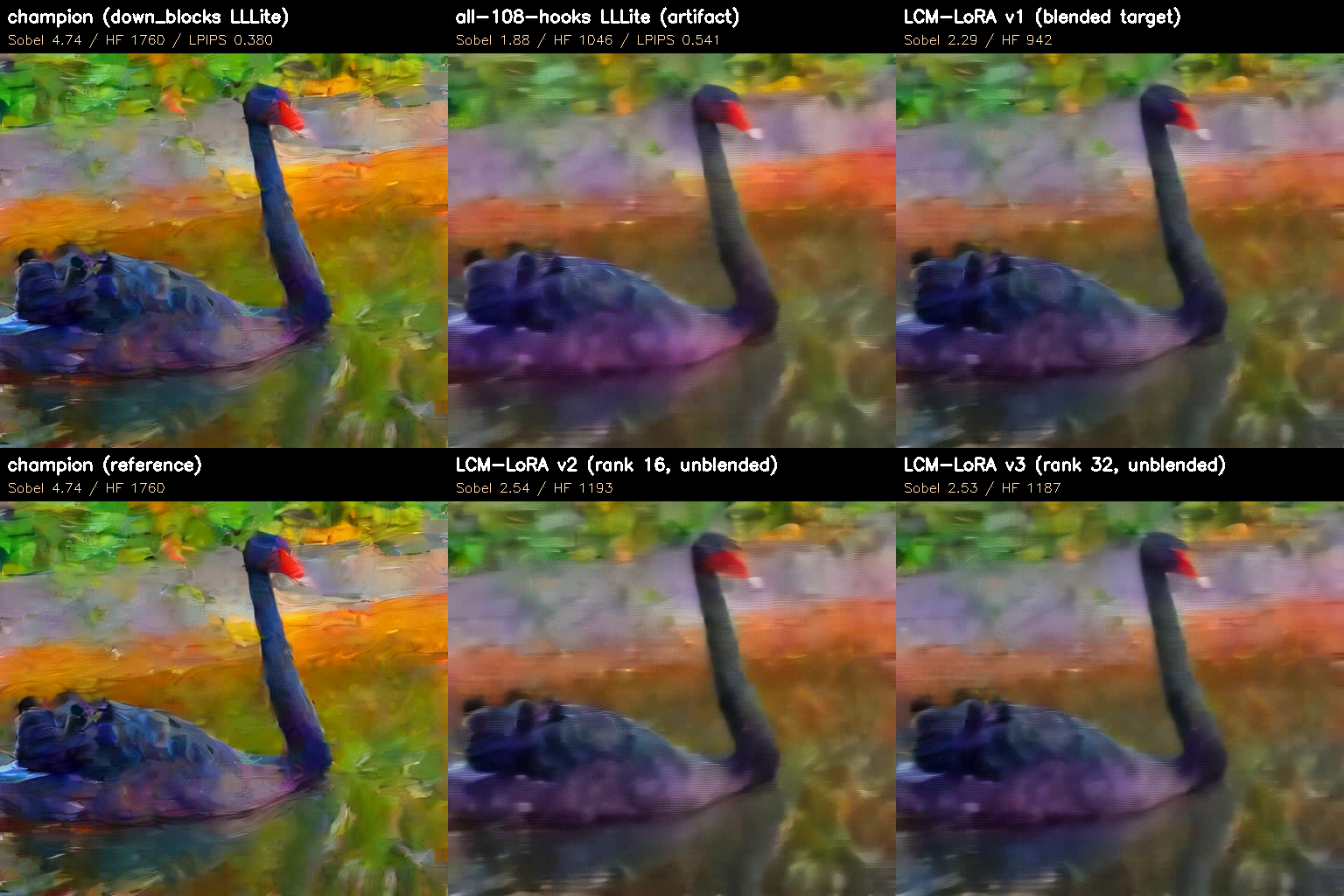}
\caption{The smoothing artifact (DAVIS \textit{blackswan}, mid-clip
frame, oil-painting prompt). \textbf{Top:} champion
(\texttt{down\_blocks} subset) preserves brushstrokes and water
reflections; all 108 LLLite hooks (middle) or a rank-16 LCM-LoRA
on blended targets (right) both collapse to a noticeably softer
character. \textbf{Bottom:} re-distilling on \emph{unblended}
teacher targets (v2, middle) recovers some sharpness; rank 32 (v3,
right) does not move the needle further.}
\label{fig:artifact}
\end{figure}

\paragraph{LCM-LoRA distillation case study: warping error rewards
smoothing} A rank-16 LoRA distilled against post-hoc-blended
($\alpha{=}0.85$) teacher targets reached $\varepsilon_w$
$10.10$ on DAVIS-9, ostensibly halving the LLLite champion's
$18.34$. Inspecting outputs revealed a $46\%$ drop in HF-FFT
($942$ vs.\ $1760$) and Sobel $2.29$ vs.\ $4.74$: the LoRA had
learned a smoothing prior, not a temporal mechanism, and
$\varepsilon_w$ rewarded the smoothness mechanically. Retraining
against unblended ($\alpha{=}1.0$) targets and doubling rank to 32
partially recovered sharpness (Sobel $2.54$, HF-FFT $1193$) but did
not close the gap. The remaining gap is a combination of (i)
$K{=}1$ inference under-sampling the diffusion trajectory of a
4-step distilled base, (ii) $v$-prediction MSE being dominated by
low-frequency error, and (iii) the choice of per-frame teacher not
transferring temporal structure that the LLLite path obtains
explicitly via the warped previous output. \emph{Methodological
takeaway:} on this pipeline, $\varepsilon_w$ alone is not a
sufficient temporal-consistency metric; pair it with a
spatial-fidelity probe (Sobel, HF-FFT, or LPIPS) before declaring
a smoothness gain. Table~\ref{tab:lcmlora} reports the full
v1 / v2 / v3 progression with the champion baseline included for
reference; Figure~\ref{fig:artifact} visualises the artifact.

\begin{table}[t]
\caption{LCM-LoRA distillation case study on DAVIS-9
(oil-painting prompt). v1 (blended teacher targets) appears to
halve $\varepsilon_w$ but does so by softening the output by
$\sim$$2\times$ on Sobel and HF-FFT. v2 (unblended targets)
recovers part of the sharpness gap; the residual is
LoRA-rank / step-count-bound, not data-bound. v3 (rank 32) does
not move the sharpness needle further. The combined v1+LLLite row
appears redundant at the metric level but inspecting the output
reveals the LLLite hooks have nothing to add to an already-flat
signal.}
\label{tab:lcmlora}
\centering
\footnotesize
\setlength{\tabcolsep}{3pt}
\resizebox{\columnwidth}{!}{%
\begin{tabular}{lr@{ }lr@{ }lrr}
\toprule
\textbf{Configuration} & \multicolumn{2}{c}{\textbf{fps}}
& \multicolumn{2}{c}{$\boldsymbol{\varepsilon_w}$}
& \textbf{Sobel} & \textbf{HF-FFT} \\
\midrule
LLLite v3 champion (no LoRA)
                            & 32.66 & $\pm$ 1.00 & 18.34 & $\pm$ 2.66 & 4.74 & 1760 \\
\midrule
v1 LoRA (blended target)    & 33.05 & $\pm$ 0.53 & 10.10 & $\pm$ 1.33 & 2.29 & 942  \\
v1 LoRA + LLLite v3         & 31.39 & $\pm$ 0.25 & 10.10 & $\pm$ 1.12 & --   & --   \\
\midrule
v2 LoRA (rank 16, unblended)& 32.15 & $\pm$ 0.82 & 10.67 & $\pm$ 1.46 & 2.54 & 1193 \\
\textbf{v3 LoRA (rank 32, unblended)}
                            & 33.02 & $\pm$ 0.60 & 10.52 & $\pm$ 1.43 & \textbf{2.53} & \textbf{1187} \\
\bottomrule
\end{tabular}%
}
\end{table}

\section{Conclusion and Future Work}\label{sec:conclusion}

We presented a systems study of video-rate streaming-throughput
stylisation on a vision-aware MLLM-conditioned distilled
edit-diffusion stack — an operating point of
$27$--$30$\,\fps{} sustained with $0.5$--$1.0$\,s end-to-end p50
latency, not an interactive low-latency one. The
inversion of the per-frame bottleneck once the U-Net is distilled —
the MLLM text encoder becomes the critical path — makes the standard
streaming-V2V design centre (denoiser acceleration) the wrong
starting point. We instead show that three engineering mechanisms —
asymmetric side-stream / main-stream pipelining with batched TE amortisation and optional static-prompt caching, a
compile-friendly LLLite reformulation, and a periodic cond-refresh
schedule with a hook subset — together sustain $27$--$30$\,\fps{}
($B\in\{8,16\}$; $32.7$\,\fps{} short-window peak) at
$512\!\times\!512$ on a single RTX~3090~Ti, $54.9$\,\fps{} on
RTX~4090, and $74.1$\,\fps{} on RTX~5090, with the released temporal
adapter transferring within in-clip noise to 19 unused DAVIS
sequences and 15 non-DAVIS clips from 7 sources. Negative results
on previous-latent mixing, cond-image manipulation, flow-warped
speculation, TensorRT FP16, and TE template pruning, together with
the LCM-LoRA smoothing case study, are intended as guardrails for
practitioners working in this regime. \emph{Limitations.} The
in-domain quality numbers use a v3 adapter trained on the same 10
DAVIS sequences they are evaluated on; the held-out splits address
this on the video axis but the held-out-prompt experiments show
adapter quality degrades gracefully outside the training-prompt
manifold. The 1-step inference recipe is empirical on \dreamlite{}
only. \emph{Future work.} Four directions are natural: an
LLLite-aware TRT~FP8 path on H100/Blackwell hardware (the
LLLite-baked export reported here is already structurally correct);
a learned 1-step distillation specialised for this pipeline with a
perceptual$+$adversarial loss (our LCM-LoRA case study suggests
$v$-prediction MSE is the wrong objective); a lower-resolution
($256^2$-native) LLLite retrain to probe whether the
asymmetric-batched-dispatch recipe still applies once the
TE / U-Net cost ratio shifts further toward the denoiser, as it
typically does on resource-constrained hardware; and a
motion-adaptive cond-refresh schedule that replaces the fixed
$N$-batch policy with a flow-magnitude or salience-driven trigger,
following the adaptive-cache lineage of
\cite{kahatapitiya2025adacache,nguyentri2025elasticcache,chen2026pfkv};
and a kernel-level swap of attention to microscaling FP4 on
Blackwell-class GPUs (SageAttention3~\cite{zhang2025sageattention3}),
which has been demonstrated lossless on diffusion stacks of
comparable scale and would compose with our compile-friendly LLLite
reformulation. Open-source code, trained adapter weights, and the
full per-sequence ablation JSONLs accompany the submission.

\section*{Acknowledgments}
The author thanks the \dreamlite{} team for releasing the base
model that this work builds on, and the kohya-ss / sd-scripts
maintainers for the LLLite reference implementation.

In accordance with the IEEE policy on the use of generative AI
(IEEE Author Center, April 2024), the author discloses that
large language model--based AI assistants (from Anthropic and OpenAI)
were used during the preparation of this work.
The AI systems were used at the following levels:
(i) \emph{writing assistance} --- copy-editing, phrasing, and
structural suggestions across all sections of the manuscript
(Abstract, Introduction, Related Work, Method, Experiments,
Negative Results, Conclusion, and this Acknowledgments section),
with all wording reviewed and edited by the author;
(ii) \emph{implementation assistance} --- code drafting,
debugging, and iterative design discussion for the training,
inference, and evaluation scripts referenced in
\S\ref{sec:method} and \S\ref{sec:exp}; and
(iii) \emph{no autonomous content generation} --- the AI systems
did not generate figures, tables, numerical results, or claims.
All research direction, experimental decisions, measurements, and
interpretive claims are the author's own, and the author takes
full responsibility for the content of the paper.

\appendices

\section{Implementation Details}
\label{app:lllite}

\subsection*{LLLite inference path}
The graph-stabilising reformulation of \S\ref{sec:method:lllite} is
realised by four inference-mode changes to the released adapter; none
alter the trained weights, and existing checkpoints load unchanged.
\emph{(i)} The cached conditioning embedding is held in a fixed-size
buffer of shape $(B_{\max},N,d_{\text{cond}})$ registered with
\texttt{persistent=False}; \texttt{set\_cond\_image} writes it
\emph{in place} so the tensor identity Dynamo closed over never
changes. \emph{(ii)} The adapter multiplier is likewise a buffer,
written via \texttt{set\_multiplier\_tensor}, so runtime strength
changes do not invalidate the graph. \emph{(iii)} A dedicated
\texttt{forward\_inference} method drops the five data-dependent
branches of the training forward (the \texttt{multiplier=0}
short-circuit, the null-conditioning guard, dtype/device checks, the
sequence-length fallthrough, and the train-only dropout path) and adds
the residual $\delta_i$ of Eq.~\eqref{eq:lllite} unconditionally.
\emph{(iv)} \texttt{apply\_lllite(\dots,inference\_mode=True)} patches
\texttt{forward\_inference} into each host Linear's \texttt{forward}
attribute in place. Together these make the LLLite-augmented forward
branch-free and identity-stable, the two conditions
\texttt{torch.compile} requires to re-use a single traced graph.

\subsection*{Off-critical-path optical flow}
The periodic conditioning refresh (\S\ref{sec:method:cond}) requires a
Farneb\"ack flow between consecutive source frames. Because
\texttt{cv2}'s Farneb\"ack implementation releases the Python GIL, we
run it across eight worker threads and dispatch it asynchronously
inside the batch-prefetch step, so the flow for batch $n{+}1$ overlaps
the compute of batch $n$ rather than stalling the per-batch wall.
This is what removes the p95 latency tail reported in
\S\ref{sec:exp:throughput}.

\subsection*{StreamDiffusion TensorRT re-run configuration}
The same-stack StreamDiffusion reference of \S\ref{sec:exp:throughput}
was produced by re-running the authors' released benchmark
(\texttt{examples/benchmark/single.py}) on an RTX~4090 host
(Ryzen~9~5900X, PyTorch~2.6.0$+$cu124, TensorRT~10.6, polygraphy 0.49).
Three compatibility patches were needed to build their engines under
TensorRT~10.x: updated \texttt{diffusers} import paths in
\texttt{tensorrt/engine.py}; a TensorRT~10 rewrite of
\texttt{allocate\_buffers} in \texttt{tensorrt/utilities.py} using
\texttt{num\_io\_tensors} and \texttt{get\_tensor\_*}; and matching the
\texttt{tensorrt-cu12} wheel to the host driver. Engine build took
$166$\,s for the U-Net (1.7\,GB FP16) and $\sim$$25$\,s each for the
tiny VAE encoder and decoder.

\subsection*{TensorRT FP16 precision attempts}
The precision-pinning remedies referenced in \S\ref{sec:neg} that
failed to recover accuracy for the wide-range MLLM TE were: per-layer
FP32 hints on the \textsc{normalization}, \textsc{softmax}, and
\textsc{matrix\_multiply} layers; the \texttt{OBEY\_PRECISION\_CONSTRAINTS}
builder flag; and an FP32-ONNX-to-FP16-engine export. The LLLite-baked
export that remained structurally correct exposed the 38 hooks'
\texttt{cond\_emb} tensors as separate engine bindings.

\bibliographystyle{IEEEtran}
\bibliography{paper}

\end{document}